\documentclass[runningheads]{llncs}
\usepackage{graphicx}
\usepackage{comment}
\usepackage{amsmath,amssymb} 
\usepackage{color}
\usepackage[dvipsnames]{xcolor}

\usepackage{subcaption}
\usepackage{cite}
\usepackage{algorithm}
\usepackage[noend]{algpseudocode}
\RequirePackage{fontawesome}

\usepackage[T1]{fontenc}

\usepackage{xspace}
\usepackage{dsfont}

\makeatletter
\DeclareRobustCommand\onedot{\futurelet\@let@token\@onedot}
\def\@onedot{\ifx\@let@token.\else.\null\fi\xspace}

\def\eg{\emph{e.g}\onedot} 
\def\ie{\emph{i.e}\onedot} 
\def\cf{\emph{c.f}\onedot} 
 
\def\wrt{w.r.t\onedot}

\def\eqwith{\quad \mbox{with}\quad}
\def\eqand{\quad \mbox{and}\quad}

\def\trans{^{\sf T}}                            
\def\transi{^{-\sf T}}                          
\def\inv{^{-1}}                                 
\def\trace{\mbox{tr}}

\newcommand{\h}[1]{{{\bf #1}}}                  
\renewcommand{\v}[1]{\mbox{\boldmath$#1$}}      
\newcommand{\m}[1]{{\tt #1}}                    
\newcommand{\mh}[1]{{^h\tt #1}}                 

\def\mcov{{\tt \Sigma}}                           
\newcommand{\s}[1]{\underline{#1}}              

\renewcommand{\vec}[0]{{\mbox{vec}}}            
\def\det{\mbox{det}}
\newcommand{\est}[1]{\widehat{#1}}              
\def\unn{\theta}

\newcommand{\mE}{{\rm I\!E}}

\def\nullspace{\mbox{\tt null}}

\newcommand{\Expectation}[0]{{\mE}}             

\def\Dispersion{\mathds{D}}                     

\newcommand{\zmatrix}[4]
  {    \left[
          \begin{array}{cc}
            {#1} \; & {#2} \\
            {#3} \; & {#4}
          \end{array}
       \right] }
\newcommand{\dmatrix}[9]
  {    \left[
          \begin{array}{ccc}
            {#1} & {#2} & {#3} \\
            {#4} & {#5} & {#6} \\
            {#7} & {#8} & {#9}
          \end{array}
       \right] }

\newcommand{\zvector}[2]
  {    \left[
          \begin{array}{c}
            {#1} \\ {#2}
        \end{array}
       \right] }
       
\newcommand{\svector}[6]
  {    \left[
          \begin{array}{c}
            {#1} \\ {#2} \\{#3} \\ {#4}\\ {#5}\\ {#6}
          \end{array}
       \right] }
       
\newcommand{\arr}[2]{\begin{array}{#1} #2\end{array}}
\newcommand{\mat}[2]{\left[\!\!\arr{#1}{#2}\!\!\right]}       

\def\gb{Gr{\"o}bner basis\xspace}

\makeatother

\begin{document}
\pagestyle{headings}
\mainmatter
\def\ECCVSubNumber{1440}  

\title{
Making Affine Correspondences Work
in Camera Geometry Computation}

\titlerunning{Making Affine Correspondences Work
in Camera Geometry Computation}
%
\author{Daniel Barath\inst{1,2} \and Michal Polic\inst{3} \and Wolfgang Förstner\inst{5} \and Torsten Sattler\inst{3,4} \and Tomas
Pajdla\inst{3} \and Zuzana Kukelova\inst{2}}
\authorrunning{Barath et al.}
%
\institute{Machine Perception Research Laboratory, SZTAKI in Budapest \and VRG, Faculty of Electrical Engineering, Czech Technical University in Prague \and Czech Institute of Informatics, Robotics and Cybernetics, CTU in Prague \and Chalmers University of Technology in Gothenburg \and Institute of Geodesy and Geoinformation, University of Bonn}
\maketitle

\begin{abstract}
Local features \eg SIFT and its affine and learned variants provide region-to-region rather than point-to-point correspondences. This has recently been exploited to create new minimal solvers for classical problems such as homography, essential and fundamental matrix estimation. 
The main advantage of such solvers is that their sample size is smaller, \eg, only two instead of four matches are required to estimate a homography. 
Works proposing such solvers often claim a significant improvement in run-time thanks to fewer RANSAC iterations. We show that this argument is not valid in practice if the solvers are used naively. 
To overcome this, we propose guidelines for effective use of region-to-region matches in the course of a full model estimation pipeline. We propose a method for refining the local feature geometries by symmetric intensity-based matching, combine uncertainty propagation inside RANSAC with preemptive model verification, show a general scheme for computing uncertainty of minimal solvers results, and adapt the sample cheirality check for homography estimation. 
Our experiments show that affine solvers can achieve accuracy comparable to point-based solvers at faster run-times when following our guidelines. We make code available at \url{https://github.com/danini/affine-correspondences-for-camera-geometry}. 
%
%
%
%
%
\end{abstract}
\section{Introduction}
\label{sec:intro}
Estimating the geometric relationship between two images, such as homography or the epipolar geometry, is a fundamental step in computer vision approaches such as Structure-from-Motion~\cite{Schoenberger2016CVPR}, Multi-View Stereo~\cite{Schoenberger2016ECCV}, and SLAM~\cite{Mur2017TRO}. 

Traditionally, geometric relations are estimated from point correspondences (PCs) between the two images~\cite{hartley2003multiple,nister2004efficient}. 
This ignores that correspondences are often rather established between image regions than between individual points to which the descriptors are finally often assigned. These regions, \ie patches extracted around keypoints found by detectors such as DoG~\cite{Lowe04IJCV} or MSER~\cite{Matas2004IMAVIS}, are oriented and have a specified size. Thus, they provide an affine transformation mapping feature regions to each other for each match~\cite{Lowe04IJCV,Philbin07CVPR}. 

Many works~\cite{Bentolila2014,PerdochMC06,Raposo2016,barath2017minimal,barath2018efficient,eichhardt2018affine,hajder2020planar,guan2020minimal,koser2009geometric,barath2017theory,pritts2019minimal} used this additional information provided by affine correspondences (ACs), \ie, region-to-region matches, to design minimal solvers for camera geometry estimation. As each correspondence carries more information, such solvers require fewer matches than their traditional point-based counterparts. For example, only three affine correspondences are required to estimate the fundamental matrix~\cite{Bentolila2014} compared to seven point correspondences~\cite{hartley2003multiple}.  This increases the probability of drawing an all-inlier sample, thus decreasing the required number of RANSAC~\cite{Fischler81CACM} iterations. 
Also, ACs are known to be more robust against view changes than point correspondences~\cite{mishkin2018repeatability}.

In terms of noise on the measurements, affine solvers are affected differently than their point-based counterparts.  If the points are well-spread in the images, the amount of noise is small compared to the distances between the points. In this case, the influence of the noise on the solution computed by a minimal solver is relatively small.  In contrast, the comparatively small regions around the keypoints that define the affine features are much more affected by the same level of noise.  As such, we observe that affine correspondence-based solvers are significantly less accurate than point correspondence-based ones if used naively. Yet, when explicitly modelling the impact of noise, we observe that affine solvers can achieve a similar level of accuracy as classical approaches while offering faster run-times. Based on our observations, we provide a practical guide for making affine correspondences work well in camera geometry computation.

\textbf{Contribution}. (\textbf{1}) We demonstrate how to use affine solvers to obtain accurate results at faster RANSAC run-times than achieved by pure point-based solvers. 
(\textbf{2}) We present strategies for all parts of the camera geometry estimation pipeline designed to improve the performance of affine solvers. This includes the refinement of the affinities defined by the features, rejection of samples based on cheirality checks, uncertainty propagation to detect and reject models that are too uncertain, and the importance of local optimization. 
(\textbf{3}) Through detailed experiments, we evaluate the impact of each strategy on the overall performance of affine solvers, both in terms of accuracy and run-time, showing that affine solvers can achieve a similar or higher accuracy than point-based approaches at faster run-times. 
These experiments validate our guidelines. 
(\textbf{4}) We make various 
technical contributions, such as a novel method for refining affine correspondence based on image intensity; a new minimal solver for fundamental matrix estimation; a strategy for combining the SPRT~\cite{chum2008optimal} test with uncertainty propagation for rejecting models early; the adaptation of the sample cheirality test, which is often used for point-based homography estimation, to affine features, and a general scheme for deriving covariance matrices for minimal solvers.
\section{Related Work}

Our guide to best use affine correspondences for camera geometry estimation problems analyzes the individual stages of the pipeline leading from matches to transformation estimates. The following reviews prior work for each stage. 

\noindent {\bf Affine features} are described by a point correspondence and a $2\times2$ linear transformation. For obtaining them, one can apply one of the traditional affine-covariant feature detectors, thoroughly surveyed in~\cite{mikolajczyk2002affine}, such as MSER, Hessian-Affine, or Harris-Affine detectors. An alternative way to acquiring affine features is via view-synthesizing, as done, \eg, by Affine-SIFT~\cite{morel2009asift}, and MODS~\cite{mishkin2015mods} or by learning-based approaches, \eg, Hes-Aff-Net~\cite{mishkin2018repeatability}, which obtains affine regions by running CNN-based shape regression on Hessian keypoints.

\noindent {\bf Matching affine regions}. Given the noise in the parameters of the regions around affine matches, a natural approach to good estimation is to use high precision least squares matching (LSM) for refining affine correspondences~\cite{lucas*81:iterative,foerstner82:geometric,ackermann84:digital}. 
Similar to template matching via cross correlation, a small patch from one image is located within a larger region  in a second image. While arbitrary geometric and radiometric models are possible, practical approaches mostly consider {\em affine transformations}. 
As a maximum likelihood estimator, LSM provides the {\em covariance matrix of the estimated parameters}, the Cramer-Rao bound, reaching standard deviations for parallaxes down to below 1/100 pixel \cite{haralick*92:computer2}. Intensity-based refinement has been used~\cite{Raposo2016,eichhardt2018affine} for pose estimation. Yet, no analysis on the accuracy of the derived uncertainty is known and a symmetric formulation of the problem is missing so far. In this paper, we close this gap by providing both. 

\noindent{\bf Affine solvers} use ACs for geometric model estimation. Bentolila and Francos~\cite{Bentolila2014} proposed a method for estimating the epipolar geometry between two views using three ACs by interpreting the problem via conic constraints. Perdoch et al.~\cite{PerdochMC06} proposed two techniques for approximating the pose based on two and three matches by converting each AC to three PCs and applying standard estimation techniques. Raposo et al.~\cite{Raposo2016} proposed a solution for essential matrix estimation from two ACs. Bar\'ath et al.~\cite{barath2017minimal,barath2018efficient} showed that the relationship between ACs and epipolar geometry is linear and geometrically interpretable. Eichhardt et al.~\cite{eichhardt2018affine} proposed a method that uses two ACs for relative pose estimation based on general central-projective views. Similarly,~\cite{hajder2020planar,guan2020minimal} proposed minimal solutions for relative pose from a single affine correspondence when the camera is mounted to a moving vehicle.
Homographies can also be estimated from two ACs as first shown by K\"{o}ser~\cite{koser2009geometric,barath2017theory}. 
Pritts et al.~\cite{pritts2019minimal} used affine features for simultaneous estimation of affine image rectification and lens distortion. 

\noindent{\bf Uncertainty analysis of image data} provides several approaches useful for our task. Variances or covariance matrices are often used to model the \emph{uncertainty of the input parameters}, \ie, image intensities, coordinates of keypoints, and affine parameters. Assuming an ideal camera with a linear transfer function, the variance of the image noise increases linearly with the intensity~\cite{dainty*74:image,szeliski10:computer}. In practice, the complexity of the internal camera processing requires an estimate of the variance function $\sigma_n^2(I)$ from the given images~\cite{foerstner00:image}. The accuracy of keypoint coordinates usually lies in the order of the rounding error, \ie, $1/\sqrt{12} \approx 0.3$ pixels. We exploit here the uncertainty of the image intensities and keypoints for deriving realistic covariance matrices of the affine correspondences. 
\begin{algorithm}
\caption{Robust model estimation pipeline with ACs}
\label{alg:pipeline}
\begin{algorithmic}[1]
  \Require{$I_1, I_2$ -- images}
  \State{$\mathcal{A} \leftarrow \texttt{DetectACs(} I_1, I_2 \texttt{)}$}\Comment{Sec.~\ref{sec:experiments:matchers}, \textit{default}: SIFT desc.~\cite{lowe1999object}, DoG shape adapt.~\cite{Lowe04IJCV} }
  \State{$\widehat{\mathcal{A}} \leftarrow \texttt{RefineACs(} \mathcal{A} \texttt{)}$}\Comment{Sec.~\ref{sec:guide:refinement}, \textit{default}: symmetric LSM refinement}
  \State{$\theta^*, q^* \leftarrow 0, 0$}\Comment{Best model and its quality}
  \While{$\neg\texttt{Terminate}$}\Comment{Robust estimation, \textit{default}: GC-RANSAC~\cite{barath2018graph}}
    \State{$S \leftarrow $ \texttt{Sample}($\widehat{\mathcal{A}}$)}\Comment{\textit{default}: PROSAC sampler~\cite{chum2005matching}}
	\If{$\neg \; $\texttt{TestSample}($S$)}\Comment{Sample degeneracy and cheirality tests, Sec.~\ref{sec:guide:chirality}}
	    \State{\texttt{continue}}
    \EndIf
    \State{$\theta \leftarrow \texttt{ModelEstimation(}S\texttt{)}$}\Comment{\textit{default}: $\m F$ -- Sec.~\ref{sec:solvers_ac}, $\m E$ -- \cite{barath2018efficient}, $\m H$ -- \cite{barath2017theory}}
	\If{$\neg \; $\texttt{TestModel}($\theta$)}\Comment{\textit{default}: tests from USAC~\cite{raguram2012usac}}
	    \State{\texttt{continue}}
    \EndIf
	\If{$\neg \; $\texttt{Preemption}($\theta$)}\Comment{Sec.~\ref{sec:propagation-minimal-solvers}, \textit{default}: SPRT~\cite{chum2008optimal} + uncertainty test}
	    \State{\texttt{continue}}
    \EndIf
    \State{$q \leftarrow \texttt{Validate(}\theta, \widehat{\mathcal{A}}\texttt{)}$}\Comment{Model quality calculation, \textit{default}: MSAC score~\cite{torr2002bayesian}}
    \If{$q > q^{*}$}
        \State{$q^{*}, \theta^{*} \leftarrow q, \theta$}
        \State{$\theta' \leftarrow \texttt{LocalOptimization(}\theta, \widehat{\mathcal{A}}\texttt{)}$ }\Comment{\textit{note}: only PCs are used from the ACs}
    	\If{$\neg \; $\texttt{TestModel}($\theta'$)}\Comment{\textit{default}: tests from USAC~\cite{raguram2012usac}}
    	    \State{\texttt{continue}}
        \EndIf
        \State{$q' \leftarrow \texttt{Validate(}\theta', \widehat{\mathcal{A}}\texttt{)}$}\Comment{Model quality calculation, \textit{default}: MSAC score~\cite{torr2002bayesian}}
        \If{$q' > q^{*}$}
            \State{$q^{*}, \theta^{*} \leftarrow q', \theta'$}
        \EndIf
    \EndIf
  \EndWhile
\end{algorithmic}
\end{algorithm}

Propagation of input uncertainty to model parameters through the estimation depends on the model being estimated. The uncertainty of a homography estimated from four or more points has been based on the SVD~\cite{criminisi01:accurate,raguram2009exploiting} and Lie groups~\cite{begelfor*05:how}. The uncertainty of an estimated fundamental matrix has been also based on the SVD~\cite{sur2008computing}, but also on minimal representations~\cite{csurka1997characterizing}. Finally, the uncertainty for essential matrices has been derived using a minimal representation~\cite{forstner2016photogrammetric}. 
As far as we know, the propagation for affine solvers has not been presented before, and there was no general scheme for deriving covariance matrices for the solutions of minimal solvers. In this paper, we provide an efficient and general scheme and the uncertainty propagation for all minimal solvers used.


\section{A Practical Guide to Using Affine Features}
As argued in Sec.~\ref{sec:intro}, and shown experimentally in Sec.~\ref{sec:experiments:matchers}, using ACs instead of point correspondences (PCs) leads to minimal solvers with smaller sample sizes but also to less accurate results. 
In the following, we analyze the individual stages of a classical matching pipeline and discuss how state-of-the-art results can be obtained with ACs. 
The pipeline is summarized in Alg.~\ref{alg:pipeline}.

\subsection{Definition of Affine Correspondences}
\label{sec:guide:definition}

Affine correspondences are defined in this paper as
\begin{equation} \label{eq:AC}
    \mbox{AC}: \quad  \{\v y_0, \v z_0, {\mh A}\}  \quad \mbox{with}
    \quad 
    \mh A =\zmatrix{\m A}{\v c}{\v 0\trans}{1}
\end{equation}
where the keypoint coordinates in the left and the right image are $\v y_0$, and $\v z_0$, and the local affinity is $\m A= \m A_2 \m A_1\inv$, \eg derived from the affine frames $\m A_1$ and $\m A_2$ representing the affine shape of the underlying image region. 
The matching and refinement of the affine region correspondences, presented in the next section, refers to the homogeneous matrix $\mh A$, specifically the translation vector $\v c$, initially  $\v 0$, and the affinity matrix $\m A$ in (\ref{eq:AC}). 

\subsection{Matching and Refining Affine Correspondences}
\label{sec:guide:refinement}
\begin{figure}[t]
  \centering	
  \includegraphics[width=0.54\textwidth]{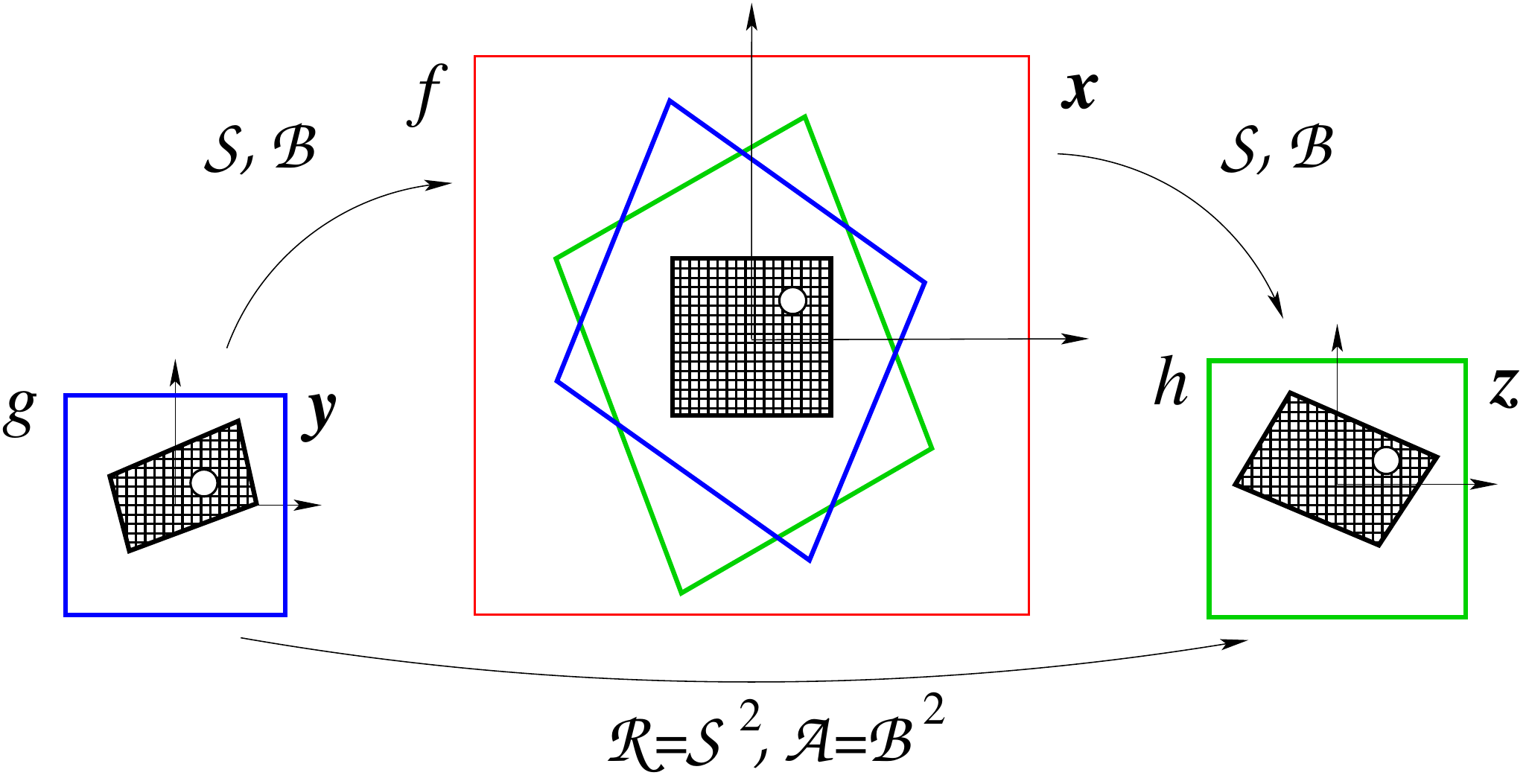}\quad
  \includegraphics[width=0.38\textwidth]{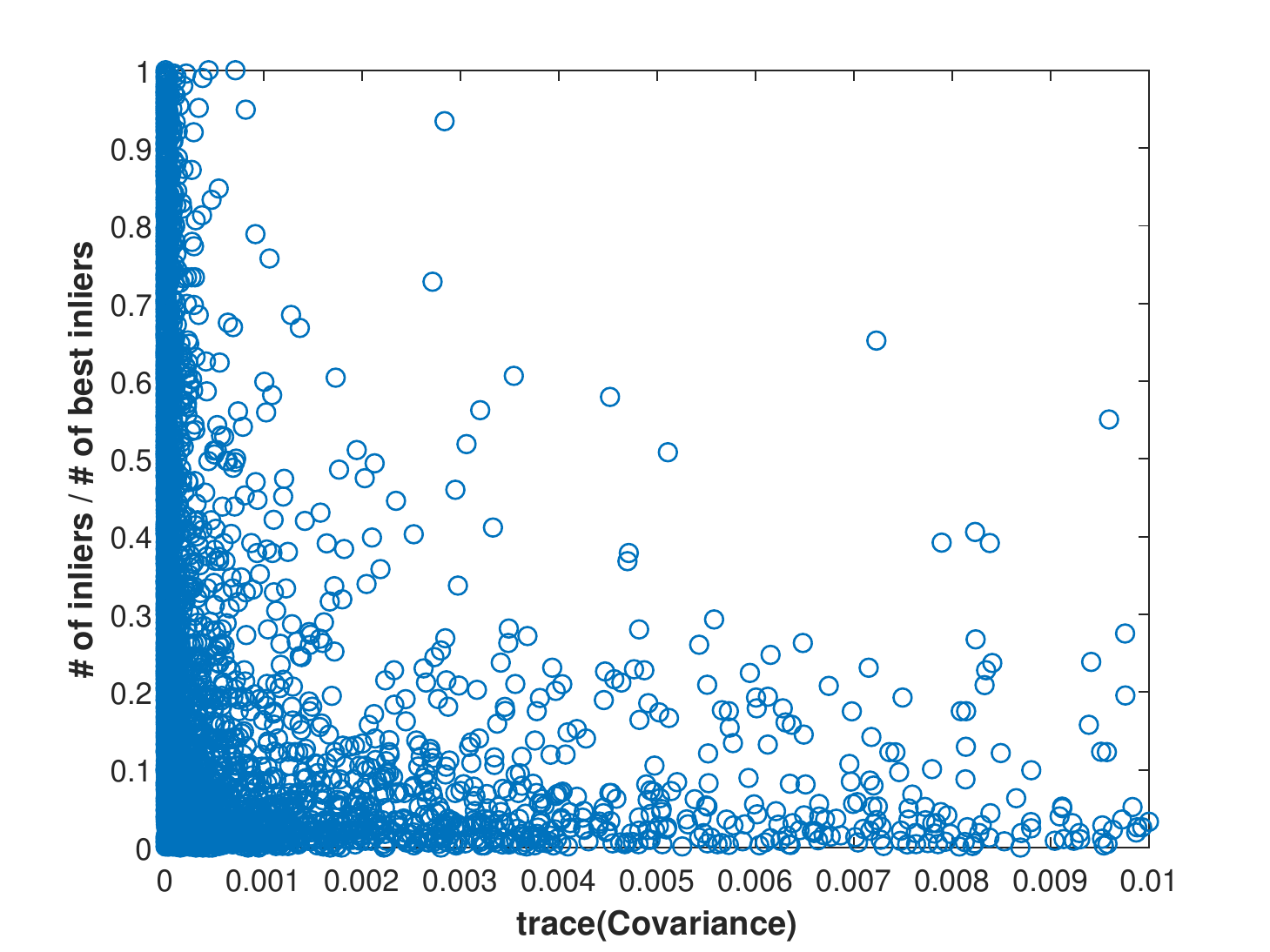}
	\caption{\textbf{Left}: Relations between two image patches $g(\v y)$ (blue) and $h(\v h)$ (green) and the mean patch $f(\v x)$ (which is the black within the red region).  The two patches $g$ and $h$ are related by geometric $\cal B$ and a radiometric $\cal S$ affinities. The correspondence is established by patch $f$, lying in the middle between $g$ and $h$. 
	We choose the maximum square (black), \ie all the pixels in $g$ and $h$ which map into the black square of the reference image $f$.
	\textbf{Right}: The inlier ratio (vertical) of 100k homographies from real scenes as a function of the trace of their covariance matrices (horizontal). This shows that uncertain models (on right) generate small numbers of inliers. We use this to reject uncertain models. 
	}
	\label{fig:LSM-matching-symmetric}
\end{figure}

For refining the ACs, we propose an intensity-based matching procedure which (i) is symmetric and provides (ii) a statistic for the coherence between the data and the model and (iii) a covariance matrix  for the estimates of the parameters. Let the two image windows in the two images be $g(\v y)$ and $h(\v z)$. We assume, both windows are noisy observations of an unknown underlying signal $f(\v x)$, with individual geometric  distortion, brightness, and contrast. We want to find the geometric distortion $\v z={\cal A}(\v y)$ and the radiometric distortion  $h={\cal R}(g)=pg+q$.  Classical matching methods assume the geometric and radiometric distortion of one of the two windows is zero, \eg $g(\v y)=f(\v x)$, with $\v y = \v x$. We break this asymmetry by placing the unknown signal $f(\v x)$ in the middle between the observed signals $g$ and $h$: $g(\v y) \; \stackrel{{\cal S}, {\cal B}}{\longrightarrow} \; f(\v x)\; \stackrel{{\cal S}, {\cal B}}{\longrightarrow} \;h(\v z)$ leading to ${\cal R} = {\cal S}^2$ and ${\cal A} = {\cal B}^2$. 
Assuming affinities for the geometric and radiometric distortions, the model is shown in Fig.~\ref{fig:LSM-matching-symmetric}(left).
The geometric and the radiometric models are
\begin{eqnarray}
		   \label{eq:2D-symmetric-affine-homogeneous}
  \v x = \m B \v y + b\,, \quad 
    \v z = \m B \v x + b \eqand
  f = s g + t\,, \quad
		   h = s f + t.
\end{eqnarray}
In the following, we collect the eight unknown parameters of the two affinities in a single vector $\v\theta = [b_{11},b_{21},b_{12},b_{22},b_1,b_2,s,t]\trans$.

Now, we assume the intensities $g$ and $h$ to be noisy with variances $\sigma_n(g)$ and $\sigma _m(h)$ depending on $g$ and $h$. Hence, in an ML-approach, we minimize the weighted sum  $\Omega(\v\unn,f)=\sum_j n_j^2(\v\theta, f)/\sigma_{n_j}^2 + \sum_k  m_k^2(\v\theta, f)/\sigma_{m_k}^2$ \wrt the unknown parameters $\v\unn$ and $f$, where the residuals are $n_j(\v\theta, f) = g_j - s\inv \left(f\left(\m B\v y_j+\v b\right)- t\right)$ and $m_k(\v\theta, f) =	h_k -\left( s f\left(\m B\inv(\v z_k-\v b)\right) + t\right)$. 
Since the number of intensities in the unknown signal $f$ is quite large, we solve this problem by fixing one group of parameters of $f$ and $\v\unn$, and solving for the other. The estimated unknown function is the weighted mean of functions $g$ and $h$ transformed into the coordinate system $\v x$ of $f$. 
The covariance matrix $\est{\mcov}_{\est\alpha\est\alpha}$ of the sought affinity is finally derived by variance propagation from ${\cal A=\cal B}^2$. 
The standard deviations of the estimated affinity $\est{\m A}$ and shift $\est{\v c}$ are below 1\% and 0.1 pixels, except for very small scales. Moreover, for the window size $M \times M$, the standard deviations decrease with on average with $M^2$ and $M$, respectively (see Supplementary Material).

\subsection{Solvers using Affine Correspondences} \label{sec:solvers_ac}
%
In this paper we consider three important camera geometry problems: estimating planar homography, and two cases of estimating relative pose of two cameras: uncalibrated and calibrated cameras.
We also include the semi-calibrated case, \ie unknown focal length, in the supplementary material.

\noindent \textbf{Homography from 1AC + 1PC:} The problem of estimating a planar homography $\m H \in \mathbb{R}^{3\times 3}$ is well-studied with simple linear solutions from point and/or affine correspondences. The homography $\m H$ has eight degrees of freedom. Since each PC gives two linear constraints on $\m H$ and each AC gives six linear constraints on $\m H$, the minimal number of correspondences necessary to estimate the unknown homography is either 4PC or 1AC+1PC. Both the well-known 4PC~\cite{hartley2003multiple} and the 1AC+1PC~\cite{koser2009geometric,barath2017theory} solvers are solving a system of eight linear equations in nine unknowns and are therefore equivalent in terms of efficiency.

\noindent \textbf{Fundamental matrix from 2AC + 1PC:}
The problem of estimating the relative pose of two uncalibrated cameras, \ie, estimating the fundamental matrix $\m F \in \mathbb{R}^{3\times 3}$, has a well-known 7PC solver~\cite{hartley2003multiple}. 
The fundamental matrix $\m F$ has seven degrees of freedom, since it is a $3 \times 3$ singular matrix, \ie, $\mbox{det}(\m F) = 0$. The well-known epipolar constraint gives one linear constraint on $\m F$ for each PC.
We propose a solver for estimating the unknown $\m F$ using the linear constraints proposed in~\cite{barath2017minimal}. 
Here, we briefly describe this new 2AC+1PC solver.

Each AC gives three linear constraints on the epipolar geometry~\cite{barath2017minimal}. 
Therefore, the minimal number of correspondences necessary to estimate the unknown $\m F$ is 2AC+1PC. The solver first uses seven linear constraints, \ie, six from two ACs and one from a PC, rewritten in a matrix form as $\m M \h f = \h 0$, where $\h f = \vec (\m F)$,
to find a 2-dimensional null-space of the matrix $\m M$. The unknown fundamental matrix is parameterized as $\m F = x\m F_1 + \m F_2$, where $\m F_1$ and $\m F_2$ are matrices created from the 2-dimensional null-space of $\m M$ and $x$ is a new unknown. This parameterization is substituted into the constraint $\det(\m F) = 0$, resulting in a polynomial of degree three in one unknown.
The final 2AC+1PC solver is performing the same operations as the 7PC solver, \ie, the computation of the null-space of a $7\times 9$ matrix and finding the roots of a univariate polynomial of degree three. 

\noindent \textbf{Essential matrix from 2ACs:}
The problem of estimating the relative pose of two calibrated cameras, \ie, estimating the unknown essential matrix $\m E \in \mathbb{R}^{3\times 3}$, has five degrees of freedom (three for rotation and two for translation) and there exists the well-known 5PC solver~\cite{nister2004efficient}.
This problem has recently been solved from two ACs~\cite{barath2018efficient,eichhardt2018affine}.
Each AC gives three linear constraints on $\m E$~\cite{barath2018efficient}. Thus, two ACs provide more constraints than degrees of freedom. One approach to solve for $\m E$ from two ACs is to use just five out of six constraints, which results in the same operations as the well-known 5PC solver~\cite{nister2004efficient} does. Another one is to solve an over-constrained system as suggested in~\cite{barath2018efficient}.
In the experiments, we used the solver of \cite{barath2018efficient} since it has lower computational complexity and similar accuracy.

\subsection{Sample Rejection via Cheirality Checks}
\label{sec:guide:chirality}

It is well-known for homography fitting that some minimal samples can be rejected without estimating the implied homography as they would lead to impossible configurations. 
Such a configuration occurs when the plane flips between the two views, \ie, the second camera sees it from the back.  
This cheirality constraint is implemented in the most popular robust approaches, \eg, USAC~\cite{raguram2012usac} and OpenCV's RANSAC.
We thus adapt this constraint via a simple strategy by converting each AC to three PCs.
Given affine correspondence $\{\v y_0, \v z_0, \m A\}$, where $\v y_0=[y_{01},y_{02}]\trans$ and $\v z_0=[z_{01},z_{02}]\trans$ are the keypoint coordinates in the left, respectively the right images,  
the generated point correspondences are $(\v y_0 + \begin{bmatrix}1, 0\end{bmatrix}\trans, \v z_0 + \m A \begin{bmatrix}1, 0\end{bmatrix}\trans)$ and $(\v y_0 + \begin{bmatrix}0, 1\end{bmatrix}\trans, \v z_0 + \m A \begin{bmatrix}0, 1\end{bmatrix}\trans)$.
When estimating the homography using the 1AC+1PC solver, the affine matrix is converted to these point correspondences and the cheirality check is applied to the four PCs.

Note that any direct conversion of ACs to (non-colinear) PCs is theoretically incorrect since the AC is a local approximation of the underlying homography~\cite{barath2017theory}.
However, it is a sufficiently good approximation for the cheirality check.

\subsection{Uncertainty-based Model Rejection} \label{sec:propagation-minimal-solvers}

Before evaluating the consensus of a model, it is reasonable to check its quality, especially to eliminate configurations close to a singularity, see \cite{frahm*06:ransac}. We can use the covariance matrix $\mcov_{\unn\unn}$ of each solution to decide on its further suitability. To do so, we propose a new general way of deriving the $\mcov_{\unn\unn}$ for minimal problems.

All problems we address here are based on a set of constraints $\v g(\v y, \v\unn)=\v 0$ on some observations $\v y$, and parameters $\v\unn$, \eg the $\m F$
estimation constraint $\v g_i(\v y_i, \v\unn)$ is of the form $\h x_i\trans\m F\h y_i=0$, hence $(\v y, \v\unn) =([\h x;\h y], \vec (\m F)) $.
We want to use classical variance propagation for implicit functions \cite{forstner2016photogrammetric}, Sec. 2.7.5. 
If $\mcov_{yy}$ is given, the determination of $\mcov_{\unn\unn}$ is based on linearizing $\v g$ at a point $(\v y,\v\unn)$ and using the Jacobians $\m A=\partial \v g/\partial \v y$ and $\m B= \partial \v g/\partial \v \unn$ leading to covariance matrix $\mcov_{\unn\unn} =  \m B\inv  \m A \mcov_{yy} \m A \trans \m B\transi$, provided $\m B$ can be inverted.
Using constraints $\v g$ we derive Jacobians $\m A$, $\m B$ algebraically. Further, given the $k^{th}$ solution of a minimal problem (a system of equations defining a minimal problem has, in general, more than one solution), \ie a pair $(\v y,\v\unn_k)$, we can compute $\mcov_{\unn_k\unn_k}$ without needing to know how the problem was solved and how this specific solution has been selected. 

However, the number of constraints in $\v g$ in most of the minimal problems is smaller than the number of parameters $\v \unn$, \eg 7 constraints vs. 9 elements of $\m F$. Then the matrix $\m B$ cannot be inverted. We propose to append the minimum number of constraints $\v h(\v\unn)=\v0$ (between the parameters only, \eg $\mbox{det}(\m F)$ and $||\m F||=1$) such that the number of all constraints $(\v g;\v h)$ is identical to the number of parameters. This leads to a regular matrix $\m B$, except for critical configurations.

Such algebraic derivations should be checked to ensure the equivalence of the algebraic and numerical solution, best by Monte Carlo simulations. However, the empirically obtained covariance matrix $\est{\mcov}_{\unn\unn}$ is regular and cannot be directly compared to the algebraically derived $\mcov_{\unn\unn}$ if it is singular (\eg for $\v \unn = \vec (\m F)$). We propose to project both matrices on the tangent space of the manifold of $\v\unn$, leading to regular covariance matrices as follows: let $\m J(\mcov)$ be an orthonormal base of the column space of some covariance matrix $\mcov$, then $\mcov_r=\m J\trans\mcov \m J$  is regular; \eg using $\m J = \nullspace(\nullspace(\mcov)\trans)$ where $\nullspace(\mcov)$ is the nullspace of $\mcov$. Hence, with $\m J=\m J(\mcov_{\unn\unn})$ the two covariance matrices $\mcov_{\unn\unn,r}=\m J\trans\mcov_{\unn\unn} \m J$ and $\est \mcov_{\unn\unn,r}=\m J\trans\est\mcov_{\unn\unn} \m J$ are regular. They can be compared and an identity test can be performed checking the hypothesis $\Expectation(\est \mcov_{\unn\unn,r})=\mcov_{\unn\unn,r}$, see \cite{forstner2016photogrammetric}, p. 71. The used constraints and detailed discussion for all listed minimal problems are in the supplementary material. 

The covariance matrix $\mcov_{\unn\unn}$ can be used in following way. Since we do not have a reference configuration, we eliminate models where the condition number $c=\mbox{cond}(\mcov_{\unn\unn})$ is too large, since configurations close to singularity show large condition numbers in practice. 
For reasons of speed, it is useful to compute an approximation for the condition number, \eg $c_s=\trace(\mcov_{\unn\unn})\trace(\mcov_{\unn\unn}\inv)$, if the inverse covariance matrix can be obtained efficiently, which is the case in our context since $\mcov_{\unn\unn}\inv = \m B (\m A\mcov_{yy}\m A\trans)\inv \m B\trans$.\footnote{For $\mcov=\mbox{Diag([a,b])}$, with $a > b$, the condition number is $c=a/b$, while the approximation is $c_s=(a+b)^2/(ab)$, which for  $a \gg b$  converges to the condition number.} A weaker measure is $\trace(\mcov_{\unn\unn})$, which is more efficient to calculate than the previous ones. It essentially measures the average variance of the parameters $\v\unn$. Thus, it can identify configurations where parameters are very uncertain. We use this measure in the following for deriving a prior for preemptive model verification by the Sequential Probability Ratio Test~\cite{chum2008optimal}.

We experimentally found, for each problem, the parameters of exponential (for points solvers) and log-normal (for affine solvers) distributions for the trace. 
These parameters are used to measure the likelihood of the model being too uncertain to lead to a large number of inliers.
In our experiments, for the sake of simplicity, we model the trace values by normal distributions for all solvers. 
Thus, we used the a-priori determined parameters to initialize the mean and standard deviation from all tested image pairs. 
Finally, we get a probability for each model being acceptable or not.  
Note that the selection of the provably correct probabilistic kernel for a particular problem and scene is a direction for future research. 
However, it is not a crucial issue due to being used only for rejecting too uncertain models early to avoid unnecessary calculations. 

As a final step, we feed the determined probability to the Sequential Probability Ratio Test (SPRT)~\cite{chum2008optimal,matas2005randomized} as a prior knowledge about the model to be verified.
This is done by initializing the model probability, which is sequentially updated in SPRT, to the a priori estimated one.


\subsection{Local Optimization}
Minimal solvers do not take noise in their input measurements into account during the estimation process. 
However, noise affects the estimated models. 
As such, not every all-inlier sample leads to the best possible transformation~\cite{chum2003locally}. 
As shown in \cite{foerstner*17efficient,schneider*17quality}, starting from the algebraic solution and performing only a single iteration of ML estimation is often sufficient to obtain a significantly better estimate. 
They show that this strategy approaches the optimal result with an error below 10\%-40\% of the parameters' standard deviations while only increasing the computation time by a factor of $\sim$2. 
A strongly recommended approach is thus to use local optimization~\cite{chum2003locally,lebeda2012fixing,barath2018graph} inside  RANSAC: 
every time a new best model is found, ML-based refinement on its inliers is used to account for noise in the input parameters. 
While this adds a computational overhead, it can be shown that this overhead is small and is compensated by the observation that local optimization (LO) typically helps RANSAC to terminate early. Moreover LO is usually applied rarely~\cite{chum2003locally}. 
As we show in Sec.~\ref{sec:experiments:lo}, local optimization is crucial to obtain accurate geometry estimates when using ACs.

\section{Experiments}
\label{sec:experiments}
In this section, different algorithmic choices are tested on homography, fundamental and essential matrix fitting problems to provide a pipeline which leads to results superior to point-based methods.

\noindent \textbf{Experimental setup.} 
Tests for epipolar geometry estimation were performed on the benchmark of~\cite{bian2019evaluation}.
The used datasets are the TUM dataset~\cite{sturm2012benchmark} consisting of videos, of resolution $640 \times 480$, of indoor scenes.
The {KITTI} dataset~\cite{geiger2012we} consists of consecutive frames of a camera mounted to a vehicle. The images are of resolution $1226 \times 370$.
Both {KITTI} and {TUM} have image pairs with short baselines.
The {Tanks and Temples} dataset~\cite{knapitsch2017tanks} provides images of real-world objects for image-based reconstruction and, thus, contains mostly wide-baseline pairs. 
The images are of sizes between $1080 \times 1920$ and $1080 \times 2048$.
The benchmark provides $1\,000$ image pairs for each dataset with ground truth epipolar geometry.
Homography estimation was tested on the scenes of the HPatches dataset~\cite{balntas2017hpatches}.
RANSAC's inlier-outlier threshold is set to 1.0 px ($\m F$), 1.0 px ($\m E$) and 5 px ($\m H$).

When evaluating $\m F$ and $\m E$ matrices, we calculate the normalized symmetric geometry errors (NSGD).
The symmetric geometry error (SGD) was proposed in \cite{zhang1998determining}.  
It generates virtual correspondences using the ground-truth ${\m F}$ and computes the epipolar distance to the estimated ${\m F}$. It then reverts their roles to compute symmetric distance. 
The SGD error (in pixels) causes comparability issues for images of different resolutions.  
Therefore, it is normalized into the range of $[0, 1]$ by regularizing by factor $f = \frac{1}{\sqrt{h^2+w^2}}$, where $h$ and $w$ are the height and width of the image, respectively.  

The error of the estimated homographies is measured by, first, projecting the first image to the second one and back to get the commonly visible area. Each pixel in the visible area is projected by the ground truth and estimated homographies and the error is calculated as the $L_2$ distance of the projected points. Finally, the error is averaged over all pixels of the visible area.


\subsection{Matchers and descriptors}
\label{sec:experiments:matchers}
\begin{figure}[h]
    \centering
	\includegraphics[width=0.95\columnwidth]{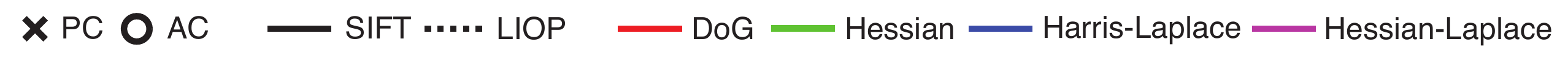}\\[2mm]
	\includegraphics[width=0.325\columnwidth]{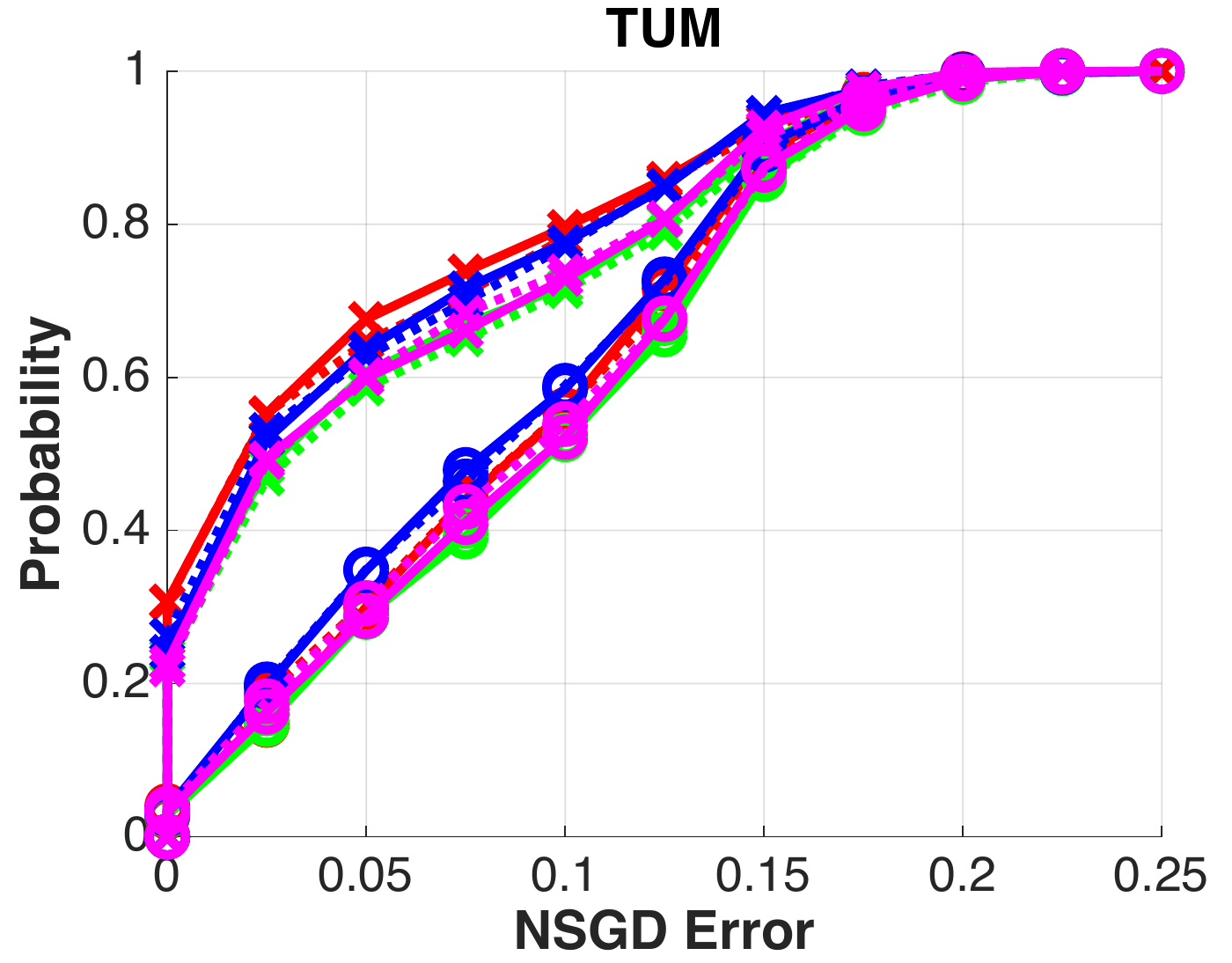}
	\includegraphics[width=0.325\columnwidth]{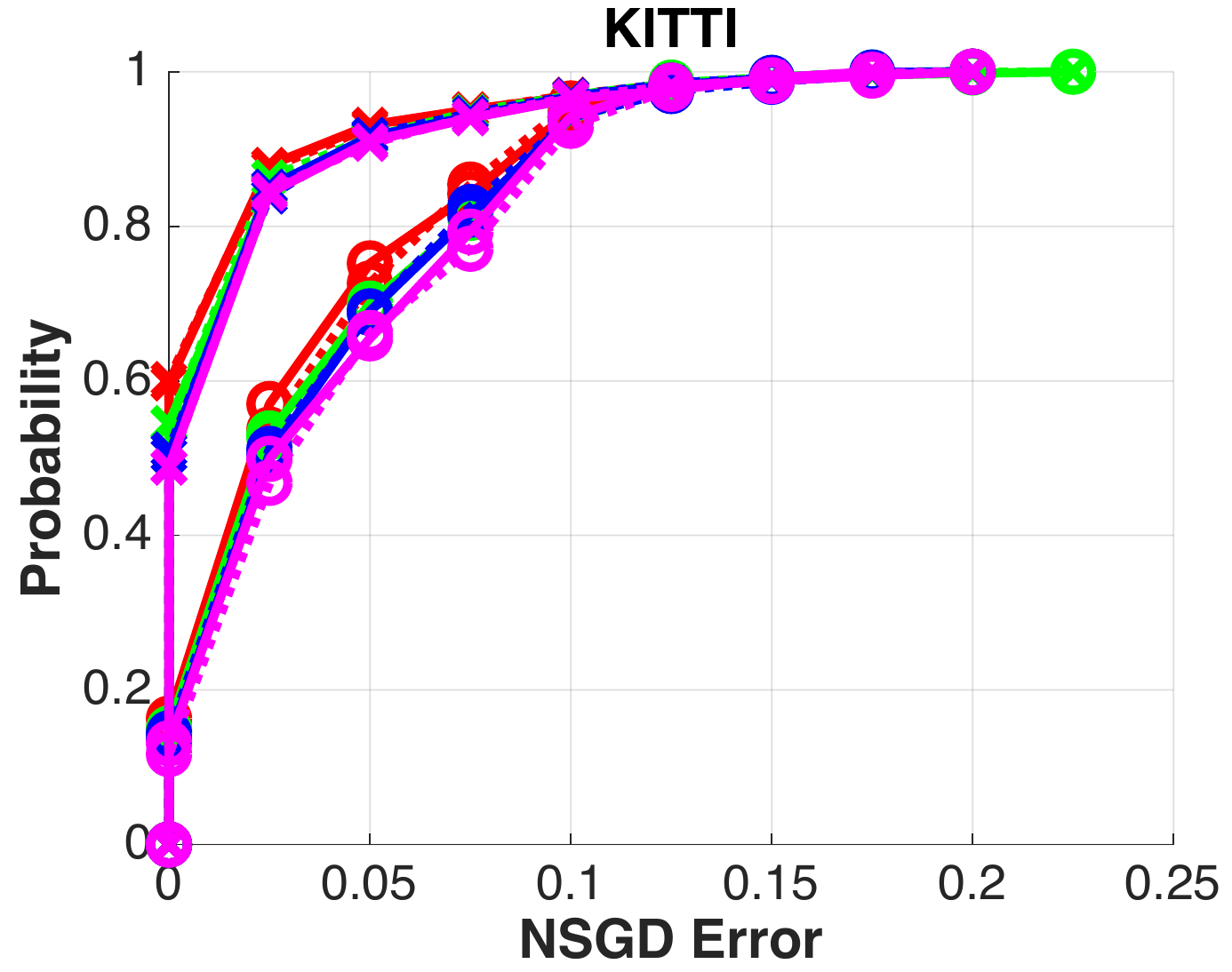}
	\includegraphics[width=0.325\columnwidth]{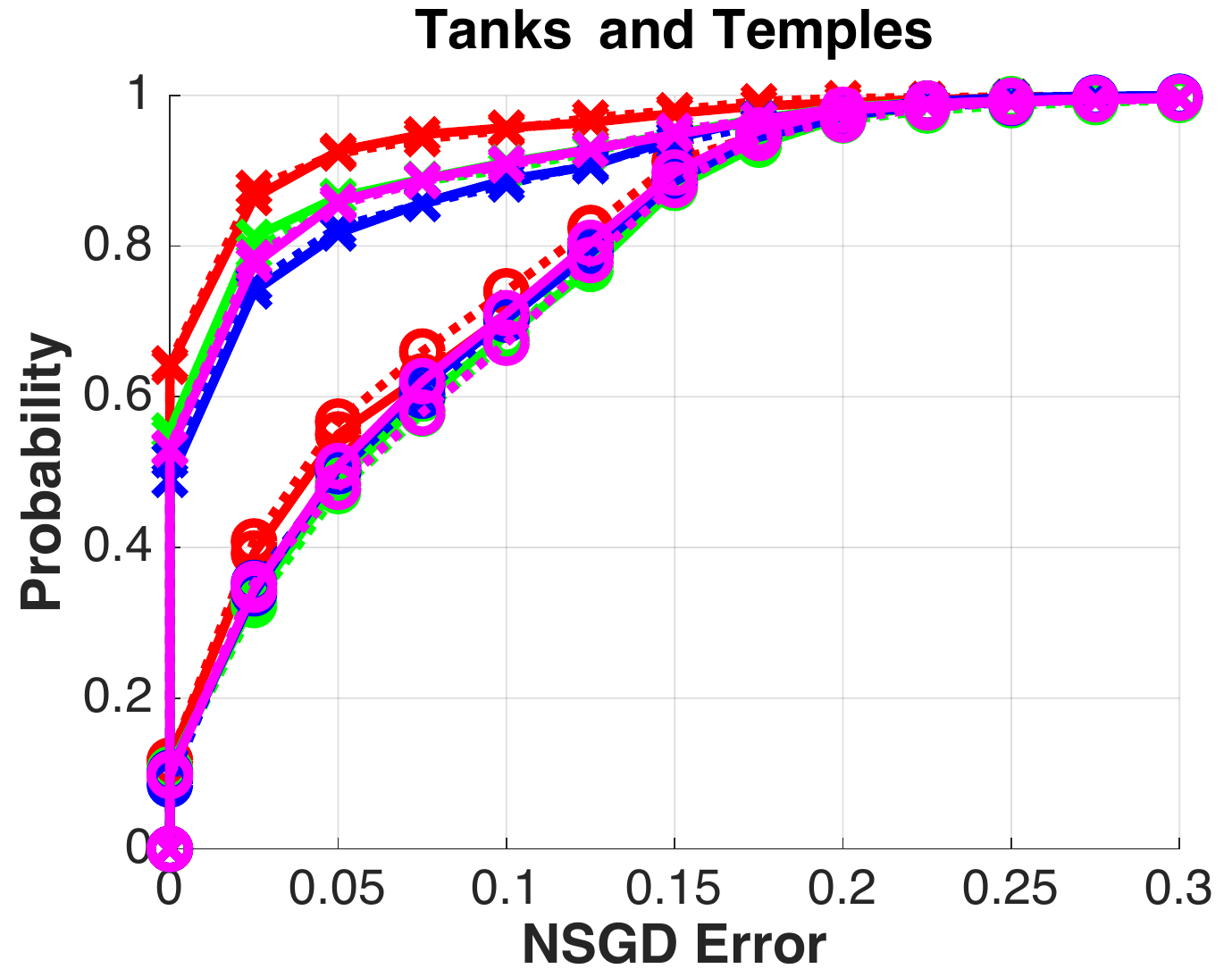}
    \caption{ Fundamental matrix estimation on datasets TUM, KITTI and Tanks and Temples (from benchmark~\cite{bian2019evaluation}; 1000 image pairs each) using ACs detected by different descriptors and detectors. 
    The CDFs of NSGD errors are shown.
    Vanilla RANSAC was applied followed by a LS fitting on all inliers.}
    \label{fig:experiment_matcher_comparison}
\end{figure}

To estimate ACs in real images, we applied the VLFeat library~\cite{vedaldi2010vlfeat} since it is available for multiple programming languages and, thus, we considered it a practical choice. 
VLFeat provides several options either for the feature descriptor or the affine shape adaptation technique.
To select the best-performing combination, first, we detected ACs using all of the possible combinations. Note that we excluded the multi-scale versions of Harris-Laplace and Hessian-Laplace~\cite{mikolajczyk2001indexing} affine shape adaptations since they were computationally expensive. 
Correspondences are filtered by the standard second nearest neighbor ratio test~\cite{lowe1999object}.
Next, we estimated fundamental matrices using affine and point-based solvers and \textit{vanilla} RANSAC~\cite{Fischler81CACM}. 
Fig.~\ref{fig:experiment_matcher_comparison} reports the cumulative distribution function (CDF) of the NSGD errors calculated from the estimated fundamental matrices. 

Curves showing affine solvers have circles as markers.
We applied PC-based methods (crosses) considering only the locations of the correspondences and ignoring the affinities.
The line style denotes the feature descriptor: straight line -- SIFT~\cite{lowe1999object}, dotted -- LIOP~\cite{wang2011local}.
Affine shape adaption techniques (DoG, Hessian~\cite{mikolajczyk2002affine}, Harris- and Hessian-Leplace~\cite{mikolajczyk2001indexing}) are shown by color. 
Applying VLFeat with any affine shape adaptation increases the extraction time by $\approx$10$\%$. 

The first and most dominant trend visible from the plots is that methods exploiting ACs are significantly less accurate then point-based approaches when the naive approach is used: vanilla RANSAC.
The SIFT descriptor~\cite{lowe1999object} and DoG affine shape adaptation lead to the most accurate results. 
Consequently, we use this combination in the experiments.
In the next sections we will show ways of making the affine solvers similarly or more accurate than point-based methods.

\subsection{Match Refinement}
We demonstrate how the proposed refinement affects the accuracy of the affine matches by analysing the fulfillment of the constraints in the case of calibrated cameras. 
For each AC, the three constraints consist of the epipolar constraint $c_p=\v c_p\trans\v e=0$ for the image coordinates, and the two constraints $\v c_a= \m C_a\trans \v e= \v 0$ for the affinity $\m A$. 
Assuming the pose, \ie the essential matrix, is known we determine a test statistics for the residuals  $c_p$ and $\v c_a$ as 
$d_p = ||c_p||_{\sigma^2_{c_p}}$ and $ d_a = ||\v c_a ||_{\mcov_{c_ac_a}}$. For the ACs of five image pairs, we used the \textsc{Lowe} keypoint coordinates and the scale and direction differences for deriving approximate affinities and refined them using the proposed LSM refinement technique (see Section~\ref{sec:guide:refinement}).

\begin{figure}
    \centering
    \includegraphics[width=0.345\textwidth]{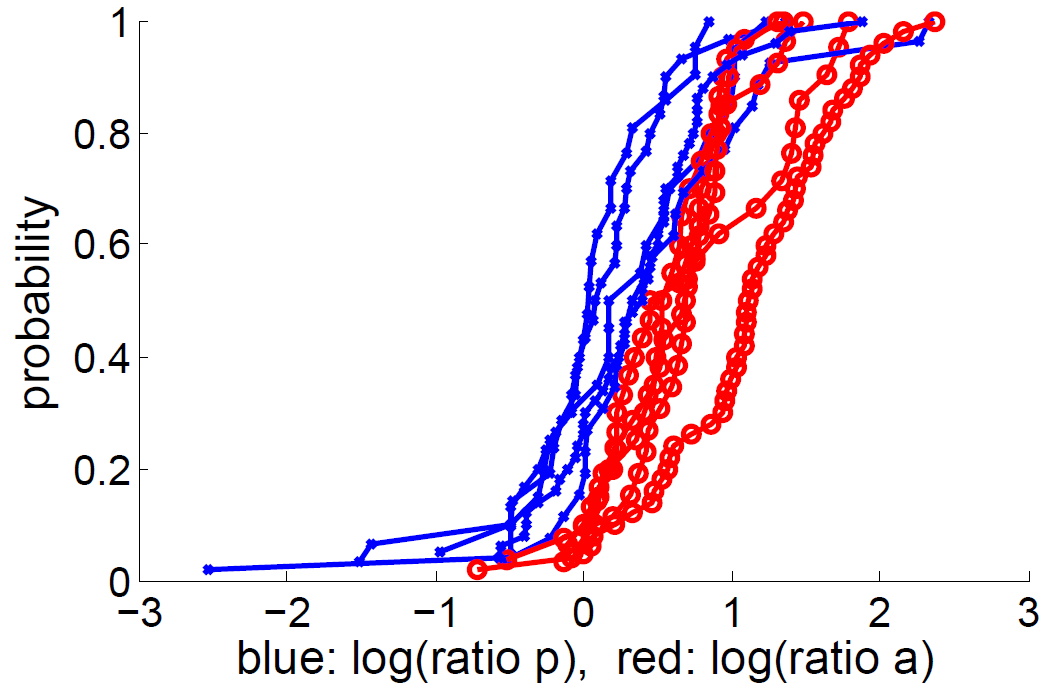}\hspace{10mm}
    \includegraphics[width=0.345\textwidth]{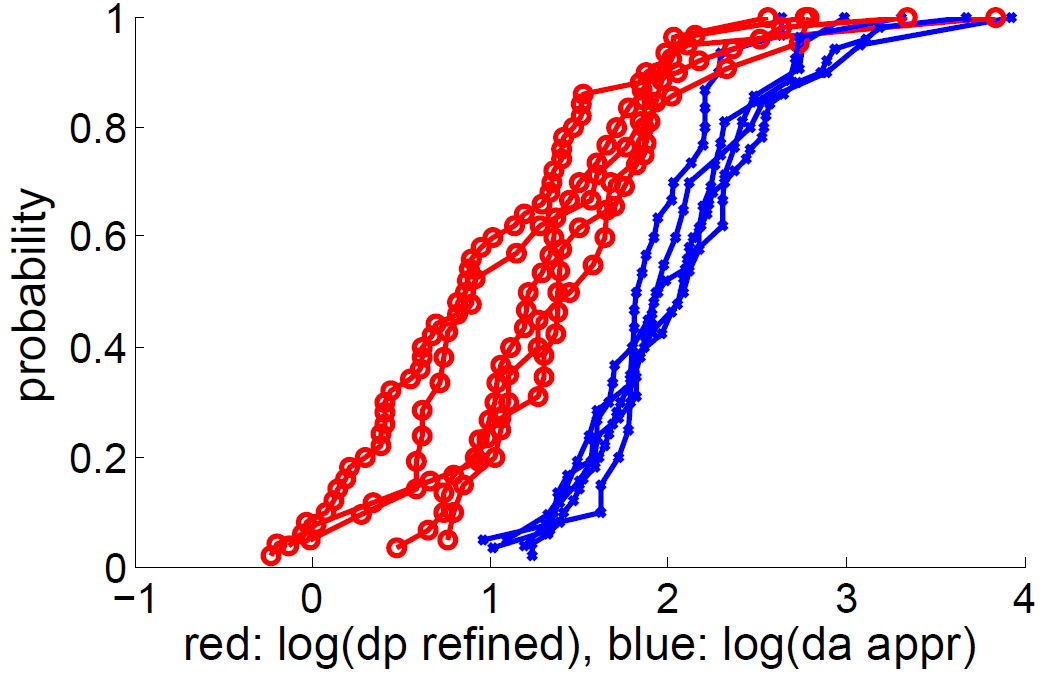}
    \caption{{\bf Left:} 
    Improvement of the points ($r_p$) and affine parameters ($r_a$) after the proposed refinement, larger values are better. 
    {\bf Right:} 
    Inconsistency with the epipolar constraints before (blue) and after (red) the refinement, smaller values are better. The CDFs are calculated from five images.}
    \label{fig:accuracy_improvement}
\end{figure}

Fig.~\ref{fig:accuracy_improvement}, left shows the CDF of the improvement caused by the proposed technique in the point coordinates ($r_p$) and in the affine parameters ($r_a$); a bigger value is better. The method improves both the affine parameters and point coordinates significantly.
In Fig.~\ref{fig:accuracy_improvement}, right the inconsistency with the epipolar constraints are shown; smaller values are better.  
The refined ACs are better, in terms of fulfilling the epipolar constraints, than the input ACs. 

\subsection{Sample Rejection via Cheirality Checks}

\begin{figure}[t]
    \centering
    \includegraphics[width=0.345\columnwidth]{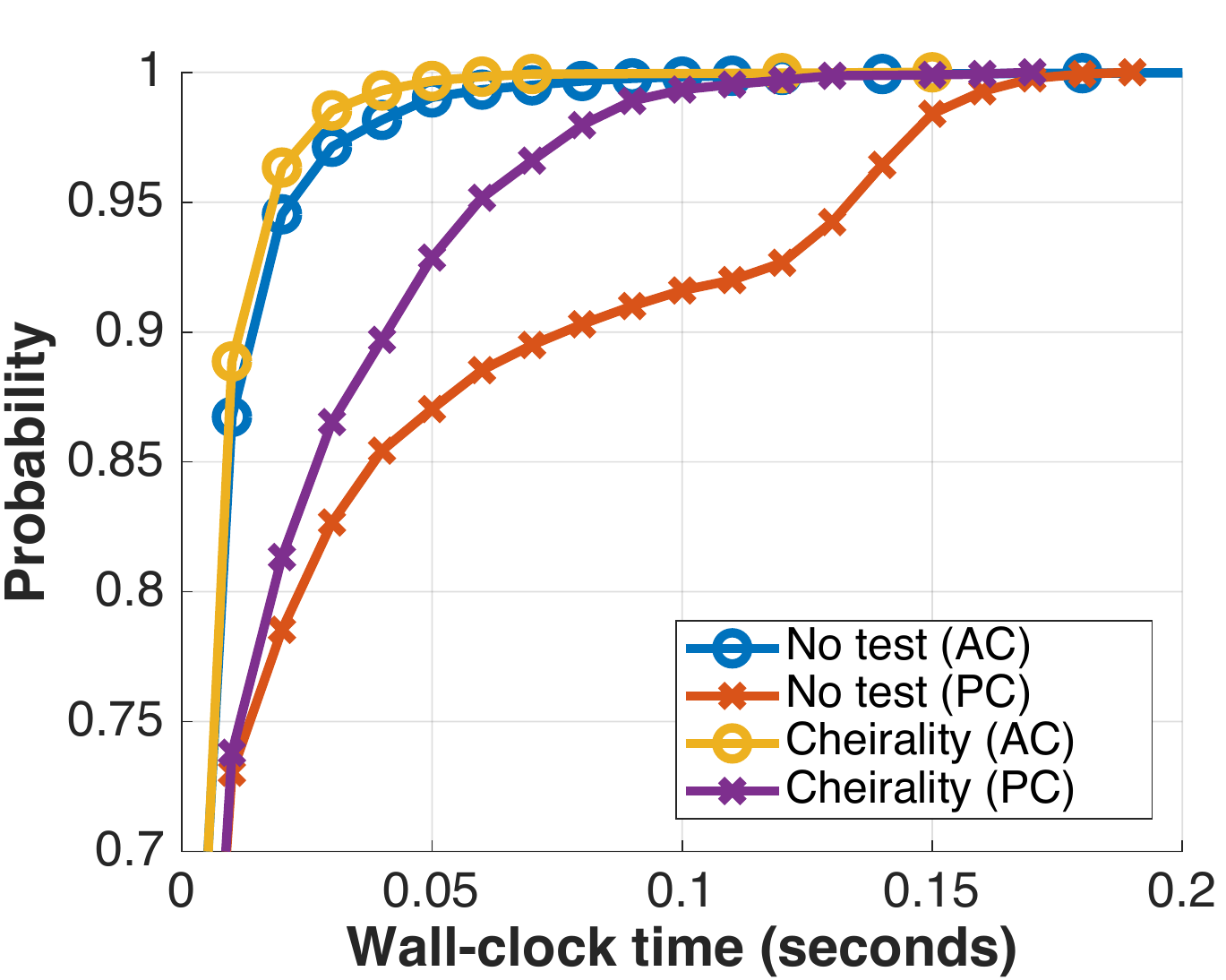}
    \includegraphics[width=0.345\columnwidth]{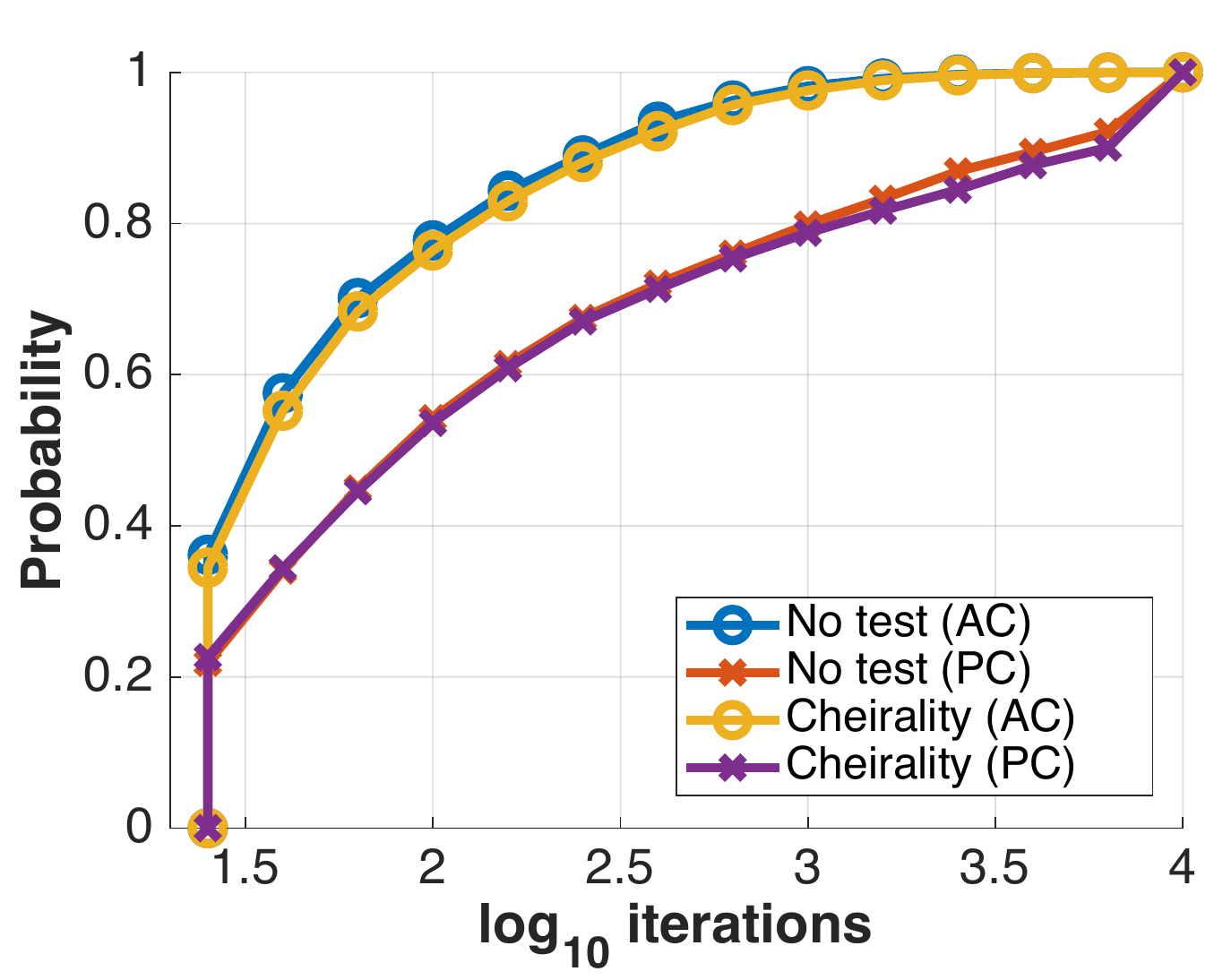}
    \caption{ Effect of cheirality test on point and affine-based homography estimators.}
    \label{fig:chirality_check}
\end{figure}

The widely used technique for rejecting minimal samples early (\ie without estimating the model parameters) when fitting homographies is the ordering check of the point correspondences as described earlier. 
Its effect when adapting it to affine correspondences is shown in Fig.~\ref{fig:chirality_check}.

In the left plot of Fig.~\ref{fig:chirality_check}, the cumulative distribution functions of the processing times (in seconds) are shown. 
It can be seen that this test has a huge effect on point-based homography estimation as it speeds up the procedure significantly.
The adapted criterion speeds up affine-based estimation as well, however, not that dramatically.  Note, that affine-based estimation is already an order of magnitude faster than point-based methods and for AC-based homography estimation the cumulative distribution curve of the processing time is already very steep, Fig.~\ref{fig:chirality_check} (left). 
This means that affine-based estimators do not perform many iterations and skipping model verification for even the half of the cases would not affect the time curve significantly. 
The avg. processing time of affine-based estimation is dropped from 8.6 to 7.7 ms.
The right plot shows the $\log_{10}$ iteration numbers of the methods.
The test does not affect the iteration number significantly.
It sometimes leads to more iterations due to not checking samples of impossible configurations, however, this is expected and does not negatively affect the time.

\subsection{Uncertainty-Based Preemptive Verification}

\begin{figure}
    \centering
	\begin{subfigure}[t]{0.325\columnwidth}
    	\includegraphics[width=1.0\columnwidth]{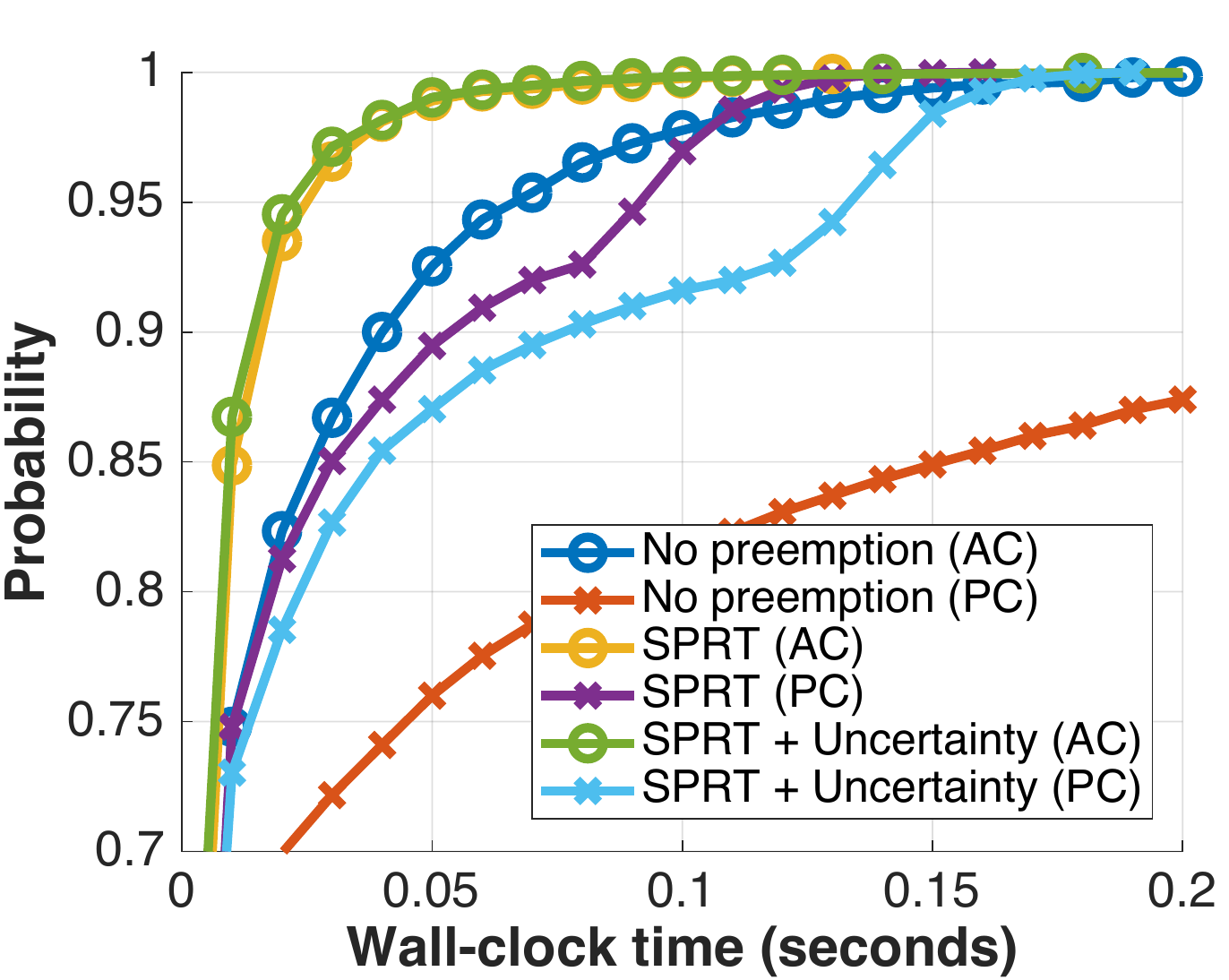}\\
		\caption{Homographies.}
	\end{subfigure}
	\begin{subfigure}[t]{0.325\columnwidth}
    	\includegraphics[width=1.0\columnwidth]{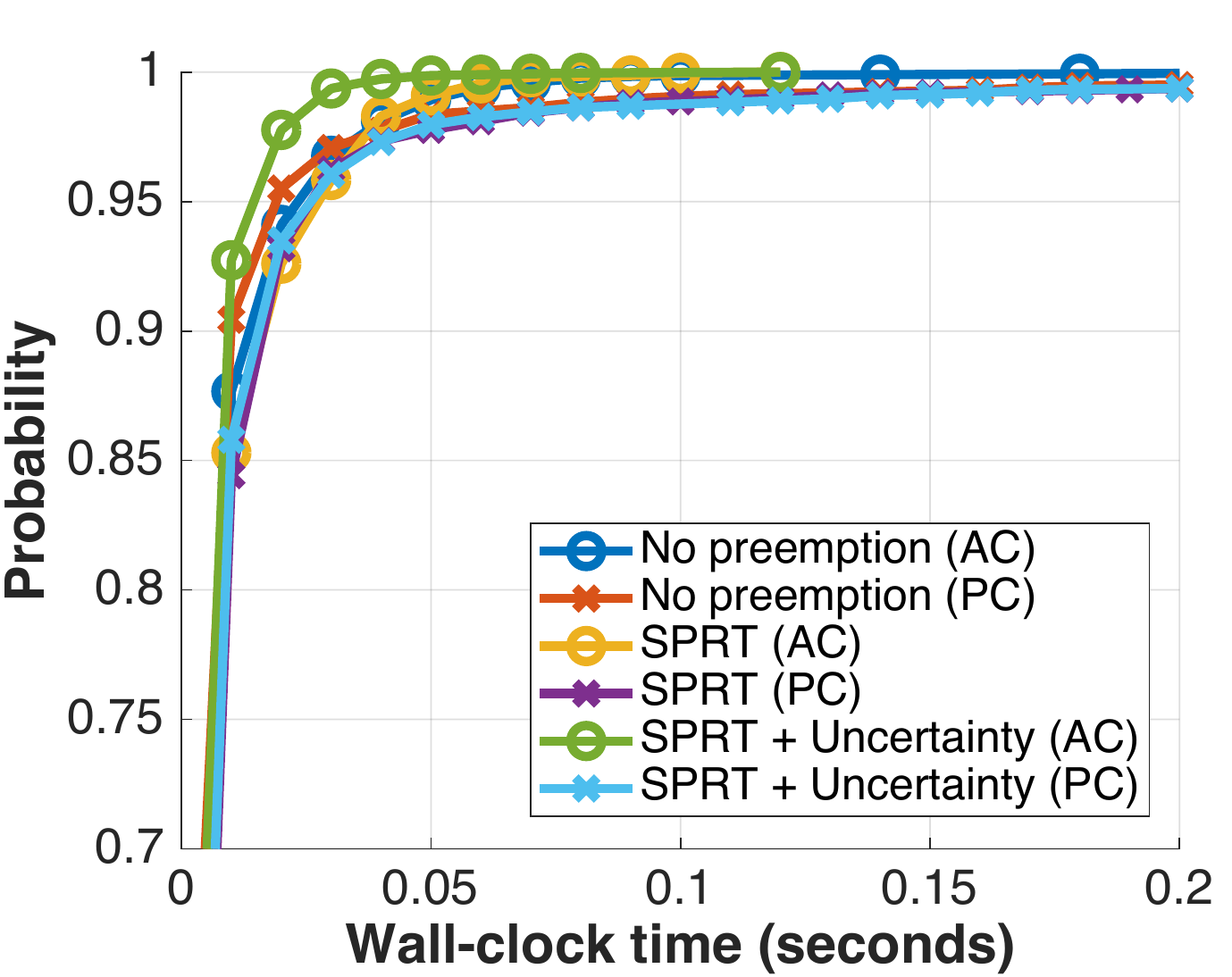}\\
		\caption{Fundamental matrices.}
	\end{subfigure}
	\begin{subfigure}[t]{0.325\columnwidth}
    	\includegraphics[width=1.0\columnwidth]{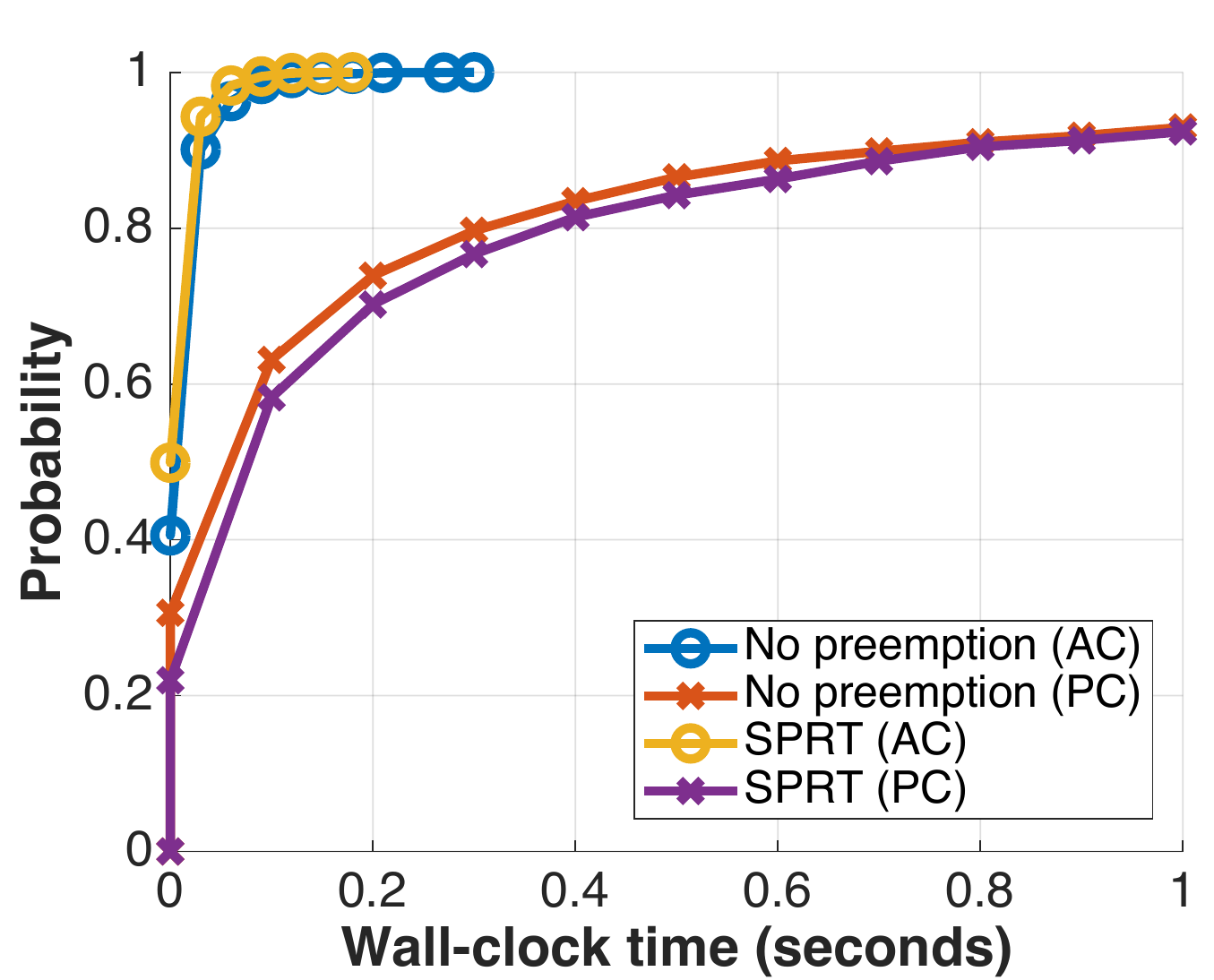}\\
		\caption{Essential matrices.}
	\end{subfigure}
    \caption{ Evaluating pre-emptive model verification strategies for affine (AC) and point-based (PC) robust estimation. The CDFs of the processing times (in seconds) are shown.
    Being fast is interpreted as a curve close to the top-left corner.}
    \label{fig:preemption_test}
\end{figure}

As it is described earlier, we combined the SPRT test~\cite{chum2008optimal,matas2005randomized} (parameters are set similarly as in the USAC~\cite{raguram2012usac} implementation) with model uncertainty calculation to avoid the expensive validation of models which are likely to be worse than the current best one. 
Fig.~\ref{fig:preemption_test} reports the CDFs of the processing time for (a) homography, (b) fundamental and (c) essential matrix fitting. 
Note that we excluded uncertainty-based verification for essential matrices since the solvers became too complex and, thus, the uncertainty calculation was slow for being applied to every estimated model.

It can clearly be seen that the proposed combination of the SPRT and the uncertainty check leads to the fastest robust estimation both for $\m H$ and $\m F$ fitting. For $\m E$ estimation from ACs, using the SPRT test is also important, leading to faster termination.
Most importantly, using affine correspondences, compared to point-based solvers, leads to a significant speed-up for all problems.

\subsection{The Importance of Local Optimization}
\label{sec:experiments:lo}
\begin{figure}
    \centering
	\begin{subfigure}[t]{0.325\columnwidth}
    	\includegraphics[trim={0.0cm 0 1.0cm 0.6cm},clip,width=1.0\columnwidth]{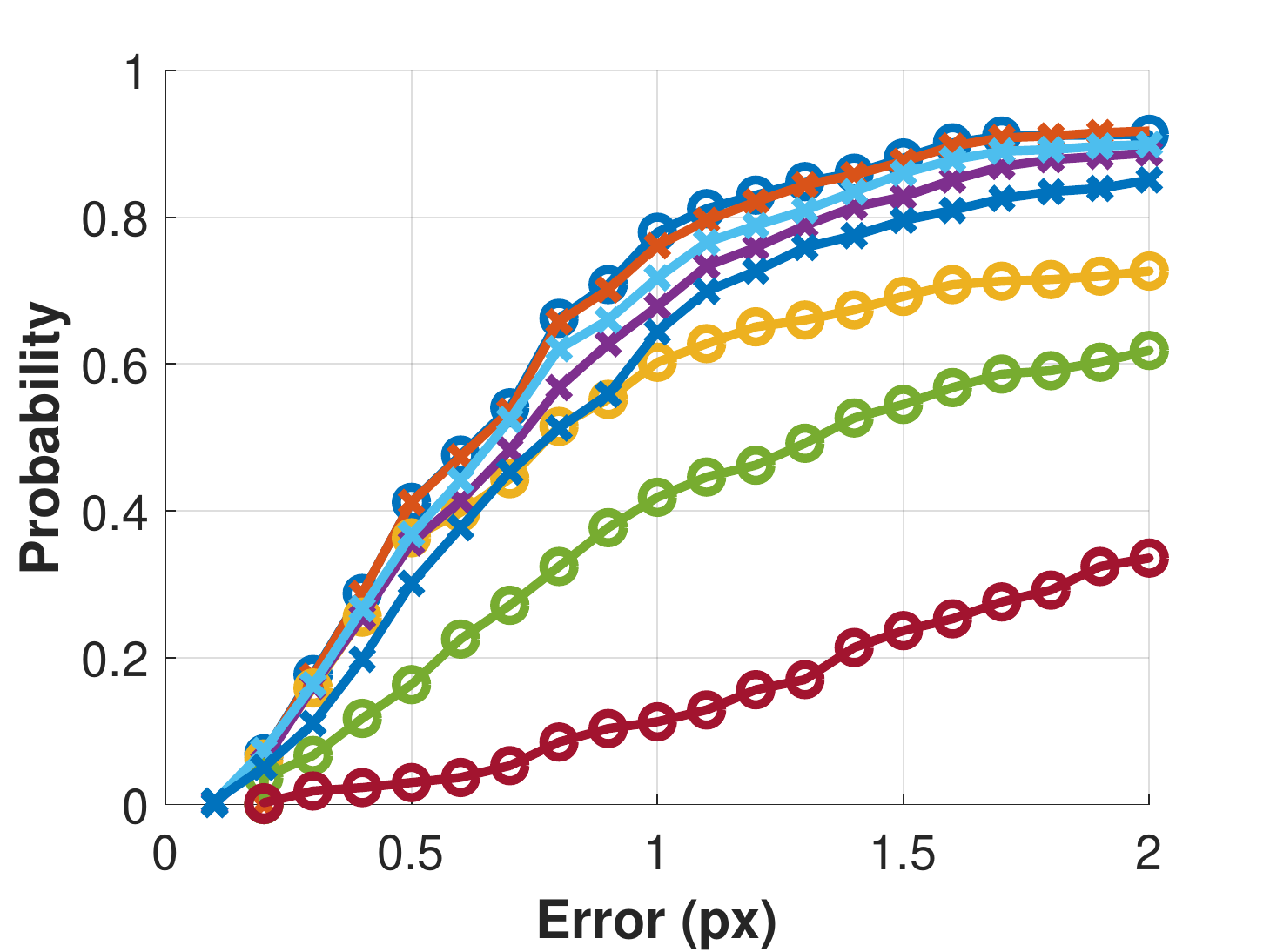}
		\caption{Homographies.}
	\end{subfigure}
	\begin{subfigure}[t]{0.325\columnwidth}
    	\includegraphics[trim={0.0cm 0 1.0cm 0.6cm},clip,width=1.0\columnwidth]{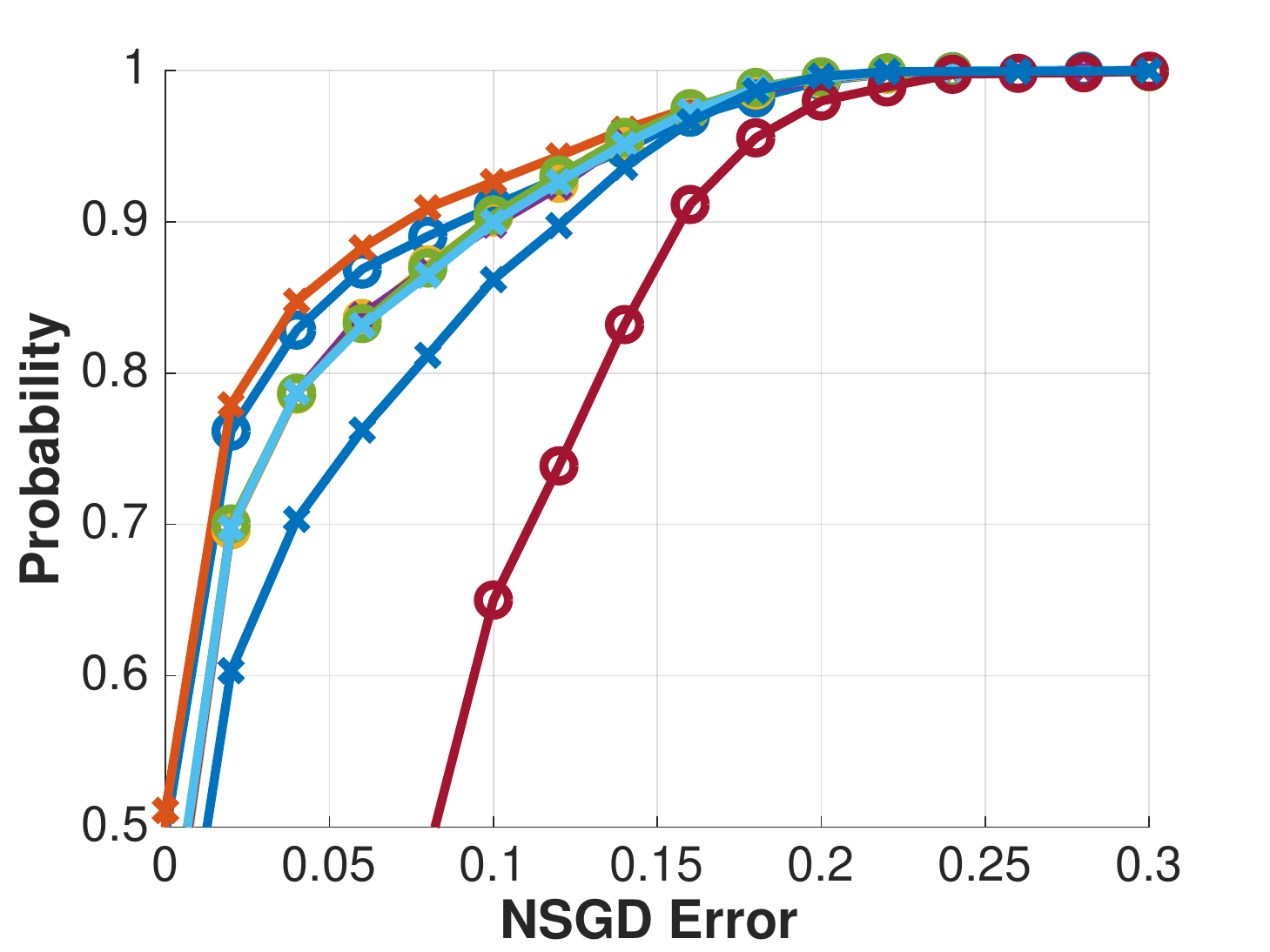}
		\caption{Fundamental matrices.}
	\end{subfigure}
	\begin{subfigure}[t]{0.325\columnwidth}
    	\includegraphics[trim={0.0cm 0 1.0cm 0.6cm},clip,width=1.0\columnwidth]{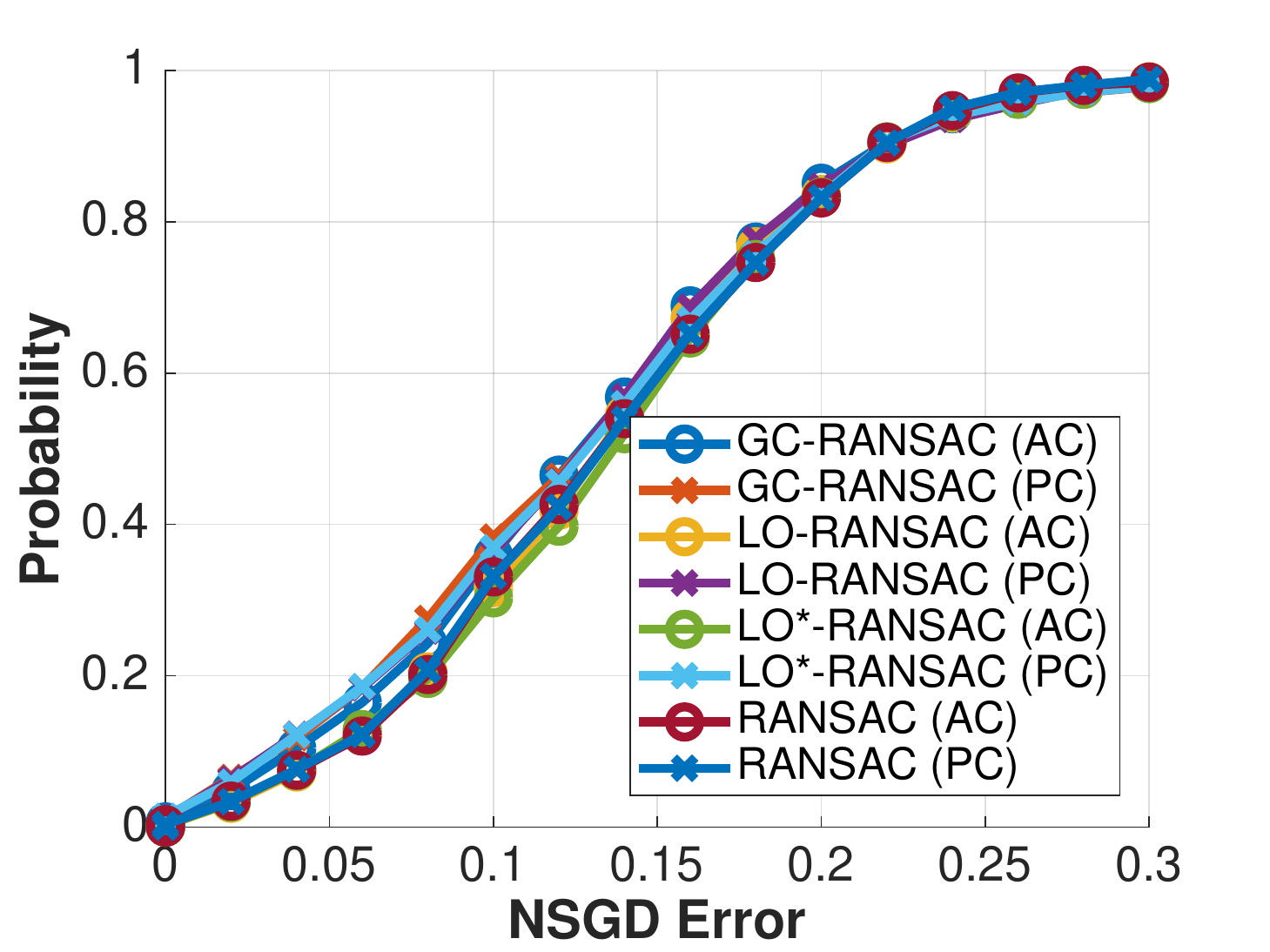}
		\caption{Essential matrices.}
	\end{subfigure}
    \caption{ Evaluating local optimization techniques for affine (AC) and point-based (PC) robust model estimation. The CDFs of the geometric errors are shown.
    Being accurate is interpreted as a curve close to the top-left corner. }
    \label{fig:local_optimization_test}
\end{figure}

While the speed-up caused by using ACs has clearly been demonstrated, the other most important aspect is to get accurate results. 
As it is shown in Fig.~\ref{fig:experiment_matcher_comparison}, including AC-based solvers in \textit{vanilla} RANSAC leads to significantly less accurate results than using PCs.
A way of making the estimation by ACs accurate is to use a locally optimized RANSAC, where the initial model is estimated by an AC-based minimal solver and the local optimization performs the model polishing solely on the inlier PCs. 
We tested state-of-the-art local optimization techniques, \ie, LO-RANSAC~\cite{chum2003locally}, LO'-RANSAC~\cite{lebeda2012fixing}, GC-RANSAC~\cite{barath2018graph}. The results are reported in Fig.~\ref{fig:local_optimization_test}. 
It can be seen that affine-based estimation with GC-RANSAC is always among the top-performing methods. In (a), it is marginally more accurate than considering only point correspondences.
In (b), using point correspondences is slighly more accurate.
Compared to the results of \textit{vanilla} RANSAC, the results of affine-based robust estimation improved notably. 

\section{Discussion}

In summary of the investigated approaches, the best practices to accurately and efficiently use affine solvers are the following.
Affinity-based model estimation has an accuracy similar or better to point-based solvers if (i) the detector and affine shape refining method is carefully selected; and (ii), most importantly, if a locally optimized RANSAC is applied to polish every new so-far-the-best model using \textit{only} the point locations from the inlier correspondences.
Consequently, affine features are used for estimating models from minimal samples, while their point counterparts are used to obtain accurate results.
Efficiency is achieved by (iii) adapting strategies well-established for point correspondences, \eg, cheirality check for homography estimation. 
(iv) Also, uncertainty-based model rejection and other preemptive verification techniques have a significant impact when speeding up the robust estimation procedure. 

\section{Conclusions}
In this paper, we have considered the problem of using affine correspondences (ACs) for camera geometry estimation, \ie, homography and epipolar geometry computation. 
Compared to classical approaches based on point correspondences (PCs), minimal solvers based on ACs offer the advantage of smaller sample sizes and, thus, the possibility to significantly accelerate RANSAC-based randomized robust estimation. 
However, noise has a larger negative impact on affine solvers as their input measurements typically originate from a smaller image region compared to point solvers. 
As we have shown, this significantly decreases the accuracy of the affine solvers. 
In this work, we have thus collected a set of ``best practices", including novel contributions such as refining the local feature geometry and uncertainty-based model rejection techniques, for using ACs in practice. 
Through extensive experiments, we have shown that following our guidelines enables affine solvers to be used effectively, resulting in similar accuracy but faster run-times compared to point-based solvers. 
We believe that our guide will be valuable for both practitioners aiming to improve the performance of their pipelines as well as researchers working on ACs as it covers a topic previously unexplored in the literature.

\subsubsection*{Acknowledgements.}

This research was supported by 
project Exploring the Mathematical Foundations of Artificial Intelligence (2018-1.2.1-NKP-00008), 
the Research Center for Informatics project CZ.02.1.01/0.0/0.0/16 019/0000765, 
the MSMT LL1901 ERC-CZ grant, 
the Swedish Foundation for Strategic Research (Semantic Mapping and Visual Navigation for Smart Robots), the  Chalmers  AI  Research  Centre  (CHAIR) (VisLo-cLearn),
the European Regional Development Fund under IMPACT No.~CZ.02.1.01/0.0/0.0/15 003/0000468, EU H2020 ARtwin No.~856994, and EU H2020 SPRING No.~871245 Projects.

\section{Supplementary Material}

This supplementary material provides the following information: 
Sec.~\ref{sec:matching} provides technical details for the matching approach described in Sec. 3.2 in the paper. 
Sec.~\ref{sec:solvers} describes a solver that estimates the essential matrix and a common focal length from affine correspondences (ACs) (\cf Sec. 3.3 in the paper). 
Sec.~\ref{sec:uncertainty} provides details on the constraints used to derive uncertainty estimates for the solvers used in the paper (\cf Sec. 3.5 in the paper). 
Sec.~\ref{sec:preemptive} discusses fitting a distribution other than the normal distribution to the trace of the covariance matrix (\cf Sec. 3.5 in the paper). 
Sec.~\ref{sec:experiments} finally provides additional results on the impact of local optimization (\cf Sec. 4.5 in the paper).

\subsection{Symmetric Least Squares Matching}
\label{sec:matching}
Here we describe the symmetric version of least squares matching in more detail and give some experimental results.
\renewcommand{\d}[1]{\mbox{\boldmath$#1$}}      
\def\unn{\theta}
\def\vunn{\v\theta}
\def\evunn{\est{\v\theta}}
\def\argmin{\mbox{argmin}}

\subsubsection{Model.}

Let the two image windows $g(\v y)$ and $h(\v z)$ in the two images be given (\cf Fig.~\ref{fig:LSM-matching-symmetric-2}). The coordinates refer to the centre of the square windows.  We assume that both windows are noisy observations of an unknown true underlying signal $f(\v x)$, with individual geometric  distortion, brightness, and contrast. We want to determine the geometric distortion $\v z={\cal A}(\v y)$ and the radiometric distortion  $h={\cal R}(g)=pg+q$.  Classical matching methods assume the geometric and radiometric distortion of one of the two windows is zero, \eg assuming $g(\v y)=f(\v x)$, with $\v y = \v x$. 
We break this asymmetry by placing the unknown signal $f(\v x)$ in the middle between the observed signals between $g$ and $h$:
\begin{equation} \label{eq:gfh}
    g(\v y) \; \stackrel{{\cal B}, {\cal S}}{\longrightarrow} \; f(\v x)\; \stackrel{{\cal B}, {\cal S}}{\longrightarrow} \;h(\v z) \quad \mbox{such that} \quad
    {\cal A} = {\cal B}^2 \;,  {\cal R} = {\cal S}^2\,.
\end{equation}
Assuming affinities for the geometric and the radiometric distortion, we have the following generative model (see Fig. \ref{fig:LSM-matching-symmetric}): 
\begin{figure}[t!]
  \centering	\includegraphics[width=0.6\textwidth]
	{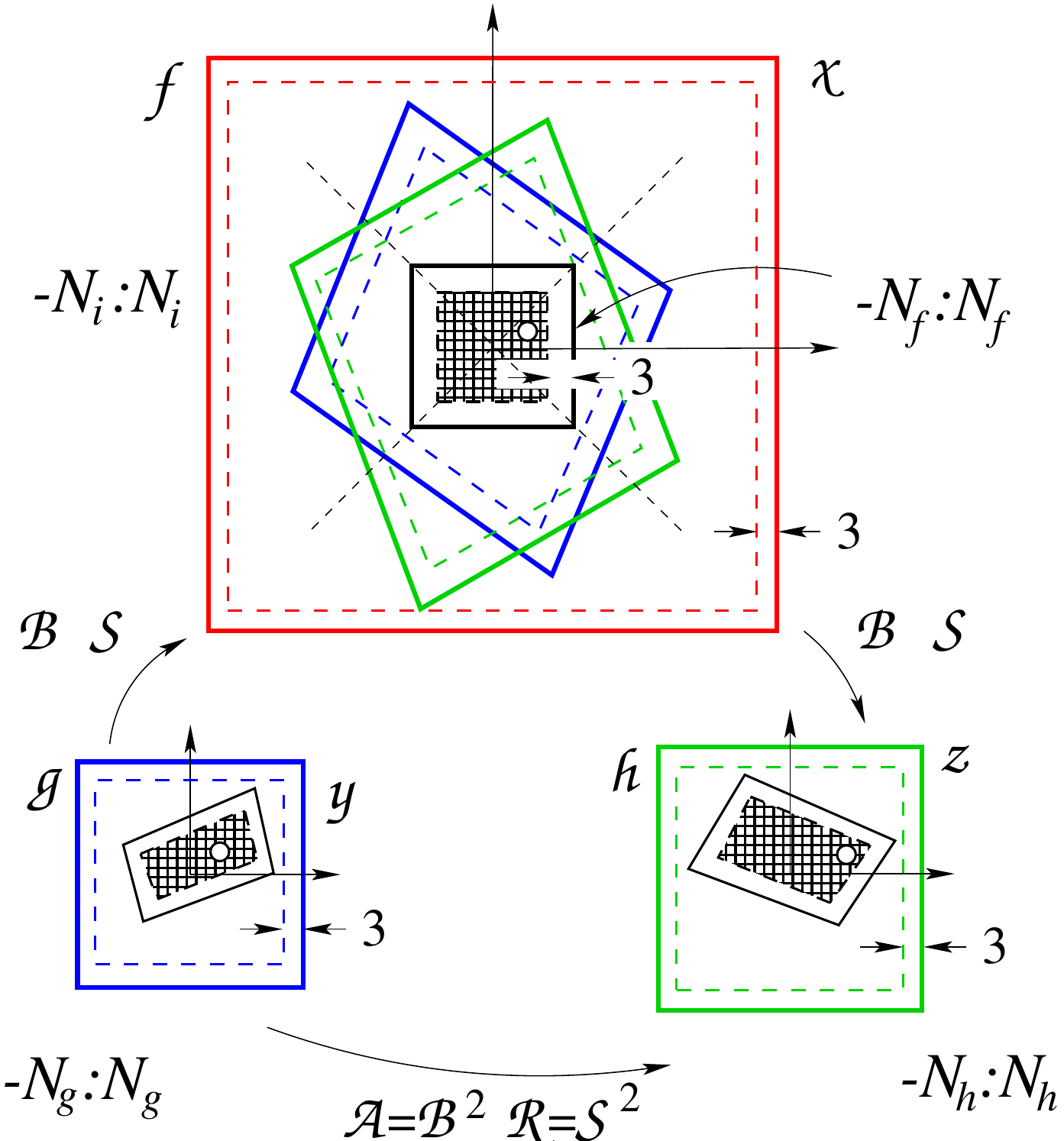}
	\caption{Relations between two given square image patches $g(\v y)$ (blue) and $h(\v h)$ (green) and the mean patch $f(\v x)$ (which is the black  within the red region).  The two image patches $g$ and $h$ are related by geometric  and a radiometric affinities $B$ and $S$, respectively. The correspondence is established by the patch $f$. Geometrically and radiometrically it lies in the middle between $g$ and $h$. Only a region in the overlap of the two patches $g$ and $h$ mapped to $f$ can be used. We choose the maximum square (black).  The observations are all pixels in $g$ and $h$ which map into the black square of the reference image $f$. We assume the reference image $f$ is a restored version of the weighted mean of the two projected images $g$ and $h$. The patches $g$ and $h$ may have different sizes. The size of the unknown signal (black, gray) depends on the sizes of $g$ and $h$, the approximate affine transformation $A$ and a border to allow bicubic interpolation, and is adapted in each iteration. The large image in the $\v x$-frame is used for generating artificial images. The dashed lines indicate the borders required for allowing bicubic interpolation}
	\label{fig:LSM-matching-symmetric-2}
\end{figure}	
The geometric and the radiometric models for the two images are
\begin{eqnarray} \label{eq:B-S}
		   \label{eq:2D-symmetric-affine-homogeneous}
    \v y \mapsto \v x: \quad \v x &=& \m B \v y + \v b \eqand 
    \v x \mapsto \v z:  \quad \v z = \m B \v x + \v b \\ \label{eq:affine-radiometry}
     g \mapsto f:\quad f &=& s g + t \eqand 
		     f \mapsto h:\quad h = s f + t\,.
\end{eqnarray}
In the following we collect the eight unknown parameters of the affinities ${\cal B}(\m B, \v b)$ and ${\cal S}(s,t)$ in the vector 
\begin{equation}
	\v\theta = \zvector{\v\theta_G }{ \v\theta_R} =\zvector{ \svector{ b_{11}}{b_{21}}{b_{12}}{b_{22}}{b_1}{b_2}} {\zvector{s}{t}}\,.
\end{equation}
 This model is rigorous only  
    in the case that the scene surface is planar in a differentiable region and the intensity differences result from brightness and contrast changes only.

We now assume the intensities $g_j$ and $h_k$ are noisy	with variances $\sigma_{n_j}^2$ and $\sigma _{h_k}^2$. These statistical properties of the noise need to be specified, \eg, assuming the variance to be signal dependent, thus \eg using $\sigma_{n_j}^{2}=\sigma_{n}^2(g(\d y_j))$ and $\sigma_{m_k}^{2}=\sigma_{m}^2(h(\d z_k))$.
When using real images, we estimate this signal-dependent variance functions of the two images, see \cite{foerstner00:image}.

Integrating the geometry and intensity transformation we arrive at the following model, which is generative, \ie, allows  to simulate observed images: 
\begin{eqnarray} \label{eq:g-function}
	\s g(\v  y_j) &=& s\inv \left(f\left(\m B\v y_j+\v b\right)- t\right)+ \s n(\d y_j)\,, \quad j=1,\ldots,J\\
	 \label{eq:h-function}
	\s h(\v  z_k) &=& \left( s f\left(\m B\inv(\v z_k-\v b)\right) + t\right)+ \s m(\d z_k)\,, \quad k=1,\ldots,K \,.
\end{eqnarray}

\subsubsection{Estimation.}

The task is to estimate the parameters $\vunn=(\vunn_G,\vunn_I)$ for the geometric and the radiometric transformation and the unknown true signal $f$ from the observed values $g(\v y_j)$ and $h(\v z_k)$.  

The explicit modeling in \eqref{eq:g-function} and \eqref{eq:h-function} allows us to write the problem as a nonlinear Gauss-Markov model with the residuals and their dispersion,	
\begin{eqnarray} \label{eq:resiuals}
	n_j(\v\theta, f) =& g_j - s\inv \left(f\left(\m B\v y_j+\v b\right)- t\right)  \hspace{2mm}\,,\quad &\Dispersion(\s n_j)=\sigma_{n_j}^2\,, \quad j=1,..., J\\
m_k(\v\theta, f) =& 	h_k -\left( s f\left(\m B\inv(\v z_k-\v b)\right) + t\right)  \,,\quad&\Dispersion(\s m_k)=\sigma_{m_k}^2\,, \quad k=1,..., K\,,
\end{eqnarray}
for all pixels $\v y_j$ of $g$  and all pixels $\v z_k$ of $h$ falling into the common region in $f$. 
Maximum likelihood (ML) estimates $(\est{\v\theta}, \est f)$ result from minimizing the weighted sum of the residuals,
\begin{eqnarray}
    \Omega(\v\theta, f)  &=& \sum_j w_j n_j^2(\v\theta, f) + \sum_k 
    w_k m_k^2(\v\theta, f)\,,
\end{eqnarray}
\wrt the unknown distortion parameters $\v\theta$ and the unknown signal $f$, using  proper weights 
\begin{equation}
	w_j=\frac1{\sigma_{n_j}^{2}} \eqand w_k=\frac 1{\sigma_{m_k}^{2}} \enspace.
\end{equation}

Due to the size of $f$, the number of unknowns is quite large. Therefore we solve this problem by alternatively fixing one group of the parameters and solving for the other: 
\begin{eqnarray}
	\evunn\mid f &=& \argmin_\unn \Omega(\v\theta, f)  \,,\\
	\est f \mid \vunn &=& \argmin_f \Omega(\vunn, f) \,.
\end{eqnarray}
Especially, the estimated unknown function is the weighted mean of the functions $g$ and $h$ transformed into the coordinate system $\v x$ of $f$, which can be calculated pixel wise:
\begin{equation}
\label{eq:f|theta}
	\est f_i \mid \evunn = \frac{^gw_{f_i} \; ^gf_i\; +\; ^hw_{f_i}\; ^hf_i}{^gw_{f_i}\; +\; ^hw_{f_i}}\,,
\end{equation}
with
\begin{equation}
	^gf_i = s \cdot g(\d y_i) +t \eqand ^hf_i = 1/s \cdot (h(\d z_i)-t)
\end{equation}
from \eqref{eq:affine-radiometry} and
\begin{equation}
	\d y_i = \m B\inv(\d x_i-\d b)) \eqand \d z_i = \m B\d x_i+ \d b
\end{equation}
The weights are
\begin{equation}
	^gw_{f_i} = \frac{1}{s^2 \cdot V_g(g(\d y_i))} \eqand ^hw_{f_i} = \frac{1}{V_g(h(\d z_i)/s^2} \,.
\end{equation}
Bicubic interpolation is used to transfer $g(\d y_i)$ and $(\d z_i)$ to $f(\d x_i)$.

As a result of the ML-estimation we obtain: (1) the parameters $\est{\v\theta}$, (2) their covariances $\mcov_{\est\unn\est\unn}$, and (3) the variance factor 
\begin{equation}
    \est{\sigma}_0^2 = \frac{\Omega(\est{\v\theta},{\est f} )}R\,,
\end{equation}
where $R$ is the redundancy of the system, \ie, the efficient number of observations $K_g+K_h$ minus the number of unknown parameters $8+K_f$, where we take the approximation $K_f = \sqrt{K_gK_h}$:
\begin{equation}
	R= K_g+K_h -(8+\sqrt{K_gK_h}) \,.
\end{equation}
If the model holds,  the variance factor is Fisher distributed with $F(R,\infty)$ and should thus be close to 1. Therefore, it is reasonable to multiply the covariance matrix $\mcov_{\est\unn\est\unn}$ with the variance factor to arrive at a realistic characterization 
\begin{equation}
	\est{\mcov}_{\est\unn\est\unn}= \est{\sigma}_0^2 \mcov_{\est\unn\est\unn}
\end{equation}
of the uncertainty of the estimated parameters. 

The covariance matrix $\est{\mcov}_{\est\psi\est\psi}$ of the parameters in the 8-vector $\v\psi$  of the geometric and radiometric affinities
\begin{equation}
    \mh A=\zmatrix{\m A}{\v a}{\v 0\trans}{1} = \dmatrix{\psi_1}{\psi_3}{\psi_5}{\psi_2}{\psi_4}{\psi_6}{0}{0}1  \eqand \mh R = \zmatrix{\psi_7}{\psi_8}{0}{1} 
\end{equation}
finally is derived by variance propagation from ${\cal A}(\d\psi)={\cal B}(\vunn)^2$,
 resulting from (\ref{eq:gfh}) and (\ref{eq:B-S}). We have
\begin{equation}
    \m A= \m B^2\,, \quad \v a = (\m B+\m I_2)\v b \eqand \psi_7 = \unn_7^2\,, \quad \psi_8 = (\unn_7+1) \unn_8\,,
\end{equation}
with the Jacobian:
 \begin{equation}
	\m J_{\psi\theta}=
	\left(\begin{array}{ccccccccc} 2\,\unn_{1} & \unn_{3} & \unn_{2} & 0 & 0 & 0 & 0 & 0 & 0\\ \unn_{2} & \unn_{1}+\unn_{4} & 0 & \unn_{2} & 0 & 0 & 0 & 0 & 0\\ \unn_{3} & 0 & \unn_{1}+\unn_{4} & \unn_{3} & 0 & 0 & 0 & 0 & 0\\ 0 & \unn_{3} & \unn_{2} & 2\,\unn_{4} & 0 & 0 & 0 & 0 & 0\\ \unn_{5} & 0 & \unn_{6} & 0 & \unn_{1}+1 & \unn_{3} & 0 & 0 & 0\\ 0 & \unn_{5} & 0 & \unn_{6} & \unn_{2} & \unn_{4}+1 & 0 & 0 & 0\\ 0 & 0 & 0 & 0 & 0 & 0 & 2\,\unn_{7} & 0 & 0\\ 0 & 0 & 0 & 0 & 0 & 0 & \unn_{8} & \unn_{7}+1 & 0 \end{array}\right)\,.
\end{equation}

Empirical tests with simulated data confirm the desired properties: (1) Exchanging the two images $g$ and $h$ leads to the inverse transformation ${\cal A}\inv$, (2) the covariance matrix derived from samples with different noise do not show significant deviations from the theoretical covariance matrix, and (3) the mean variance factor is not much larger than 1, usually by 20\% to 40\%. The deviations from 1 are significant and can be explained by the approximations resulting from the bicubic interpolations of the same function based on different grids. Hence the internal quality measures can be used for self-diagnosis.

\begin{figure}[t]
    \centering
    \includegraphics[width=0.24\textwidth]{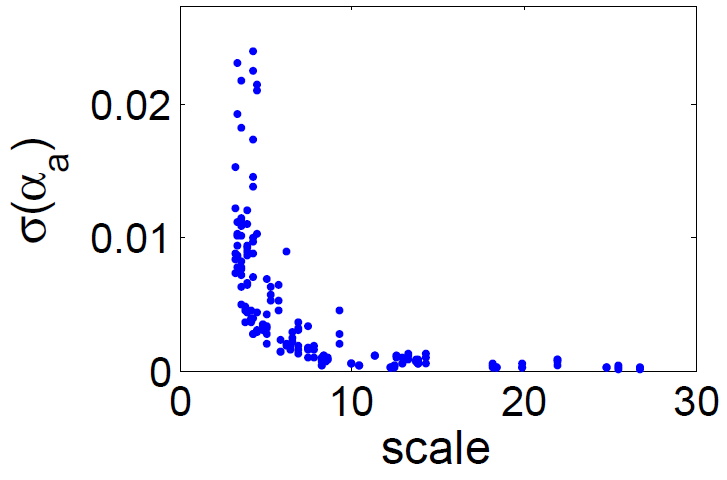}
    \includegraphics[width=0.24\textwidth]{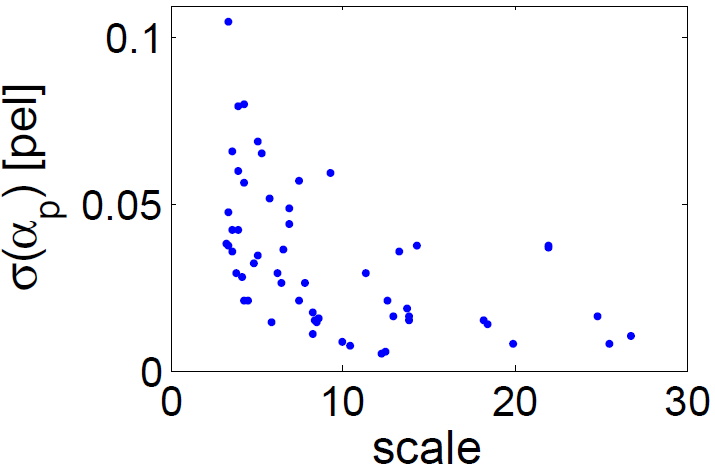}
    \includegraphics[width=0.24\textwidth]{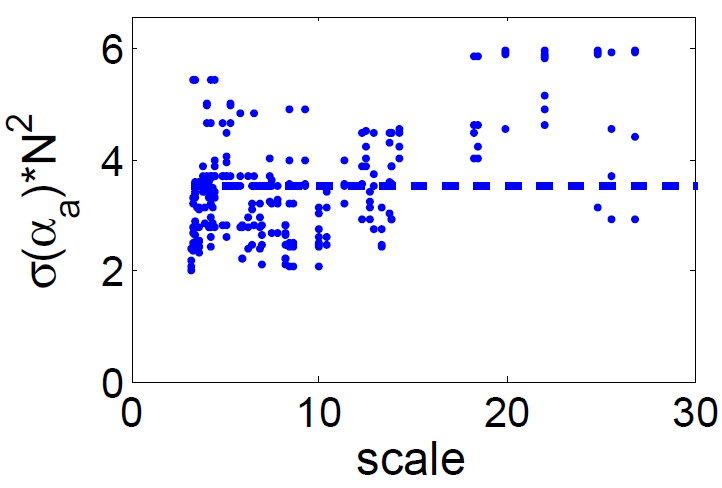}
    \includegraphics[width=0.24\textwidth]{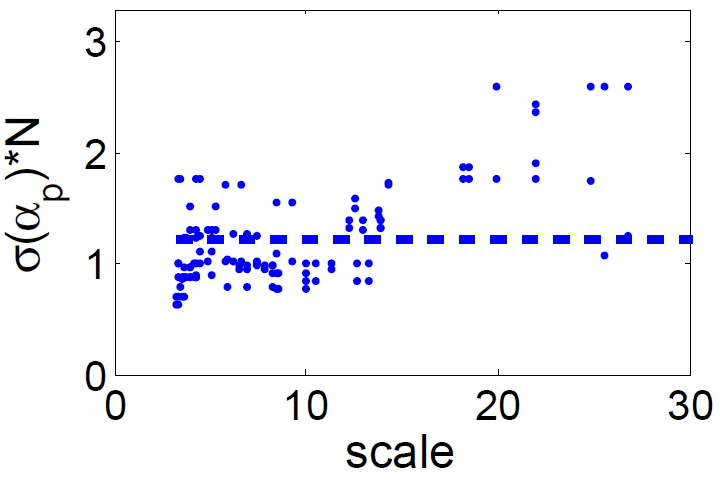}
    \caption{Predicted standard deviations of affine parameters as a function of the \textsc{Lowe}-scale of the affine correspondence. {\bf 1 and 2:} Non-normalized values  $\sigma_{\alpha_i},i=1,...,4$  are unit less, the values  $\sigma_{\alpha_i},i=5,6$ have unit [pixel]. Observe, subpixel accuracy can be reached and the affinity parameters have a standard deviation below 1\%, except for very small scales. {\bf 3 and 4:} Normalized standard deviations $N^2\,\sigma_{\alpha_i},i=1,...,4$ and $ N\sigma_{\alpha_i} ,i=5,6$ would be constant if the texture in all windows had the same gradient variance. The deviations from a constant reflect the variation of the texture in the images. Observe, the ratio between the mean values for the normalized standard deviations of the affine and the shift parameters is 3.7/1.3=2.77 (see the dashed lines in 3 and 4). It is in fair coherence with the theoretical value $\sqrt{12}=3.46$} 
    \label{fig:predicted_standard_deviation}
\end{figure}

We now show the potential of the refinement of the affinitiy using LSM, namely: the expected precision of the affinity  for ideal cases.
This is based on the part of the normal equation matrix $\m N$ related to the 6 parameters of the geometric affinity: $\m N = \sigma_n^{-2}\sum_{ij}\nabla f_\unn(i,j)\nabla f\trans_\unn(i,j)$, where the sum is over all pixels in an $N\times N$ window. If we assume the distortion is zero and  $f$ is known, then the gradient is $\nabla f = [xf_x,yf_x,xf_y,yf_y,f_x,f_y]$. 
Observe, the $2\times2$ matrix referring to the translation parameters is proportional to the structure tensor of the patch. 
We now assume that the gradients in the window have the same variance $\sigma_{f'}^2$ and are mutually uncorrelated. Then the normal equation matrix will be diagonal leading to the covariance matrix
\begin{equation}
    \mcov_{\alpha\alpha} = \zmatrix{\sigma_a^2\m I_4}{\m 0}{\m 0}{\sigma_p^2\m I_3}  \eqwith \sigma_a= \frac{\sqrt{12}}{N^2} \frac{\sigma_n}{\sigma_{f'}} \eqand \sigma_p= \frac{1}{N}\frac{\sigma_n}{\sigma_{f'}} \,.
\end{equation}
Fig. \ref{fig:predicted_standard_deviation} shows the predicted standard deviations for real image patches having different sizes, depending on the \textsc{Lowe}-scale $s$ using $M=7s$. In spite of the variation of the texture within the individual patches, the theoretical relation to the window size is visible. Deviations for large scales result from the fact that sometimes the interior of such patches is not highly textured. 

\subsection{Solvers using ACs -- Semi-calibrated case}
\label{sec:solvers}

In this section, the solver for AC-based relative pose and focal length estimation is discussed.

\noindent{\textbf Essential matrix and focal length from 2ACs}: 
The problem of estimating the relative pose and a common focal length of two semi-calibrated cameras, \ie, estimating the unknown essential matrix $\m E \in \mathbb{R}^{3\times 3}$ and focal length $f \in \mathbb{R}$, has six degrees of freedom (three for rotation, two for translation and one for focal length) and there exists several well-known hidden variable or \gb 6PCs solvers~\cite{hartley2012efficient,stewenius-kahl-etal-ccvpr-05}.
This problem has recently been solved from 2ACs~\cite{barath2017minimal}.
Each AC gives three linear constraints~\cite{barath2018efficient}. 
One approach to solve for $\m E$ and $f$ from 2ACs is to apply the solver of~\cite{barath2017minimal}. The 2ACs solver~\cite{barath2017minimal} as well as the hidden variable and \gb 6PCs solvers~\cite{hartley2012efficient,stewenius-kahl-etal-ccvpr-05} are based on ten well-known constraints on the essential matrix $\m E$, \ie, the singularity constraint $\det(\m E) = 0$ and the trace constraint $2\m E\m E^\top \m E - \mbox{tr}(\m E\m E^\top)\m E = \m 0$. These solvers are therefore solving a system of ten polynomial equations in three unknowns.

Recently, however, a more efficient solver for the 6PC problem was proposed in~\cite{kukelova2017clever}. 
This solver is based on the fact that for a fundamental matrix of the form  $\m F = \m K^{-\top} \m E \m K^{-1}$, where $\m K = \mbox{Diag}([f,f,1])$ is the calibration matrix with focal length $f$, two constraints can be derived. 
These constraints can by obtained by eliminating the focal length from the singularity and the trace constraint. The first constraint corresponds to the singularity constraint and the second one is a fifth degree polynomial in the elements of $\m F$ (see Eq 18 in~\cite{kukelova2017clever}). The final solver, after calculating the three dimensional null-space using six equations from the epipolar constraint, then solves these two equations in two unknowns
(unknowns from the parameterization of $\m F$ as a linear combination of basis null-space vectors). 
Note that in the solver the scale of $\m F$ is fixed by fixing one element in the linear combination of the null-space vectors. 
However, for the uncertainty propagation the scale will be fixed by applying a different constraint on $\m F$, \ie, the constraint $||\vec (\m F)|| - 1 = 0$.
The parameterization from~\cite{kukelova2017clever}, with two constraints instead of ten from the singularity and the trace constraints~\cite{stewenius-kahl-etal-ccvpr-05,hartley2012efficient}, not only leads to a simpler solver but will also be useful for fast uncertainty propagation as we will discuss in Sec.~\ref{sec:propagation-Ef}.

The formulation with two constraints on $\m F$ can be straightforwardly adapted for the 2ACs solver. 
We propose to use the solver from~\cite{kukelova2017clever} by first  calculating the null-space using the linear system which 2 ACs imply. The rest of the solver remains unchanged and can be similarly applied as when using point correspondences.

\subsection{Uncertainty Calculation}
\label{sec:uncertainty}

All geometric problems we address in the paper are based on a set of constraints $\v g(\v y, \v\unn)=\v 0$ between some observations $\v y$, and some parameters $\v\unn$. The classical variance propagation for implicit functions leads to
\begin{equation}
    \mcov_{\unn\unn} =  \m B\inv  \m A \mcov_{yy} \m A \trans \m B\transi \eqwith 
    \m A=\frac{\partial \v g}{\partial \v y}   \eqand    \m B=\frac{\partial \v g}{\partial \v \unn}\,,
\end{equation}
see \cite[Sect. 2.7.5]{forstner2016photogrammetric}. In our context, the covariance matrix $\mcov_{yy}$ refers to the input measurements (\eg, keypoints coordinates or affinity correspondences), the covariance matrix $\mcov_{\unn\unn}$ refers to the model parameters.

For minimal problems, the number of constraints $\v g$ is usually smaller than the number of parameters of the model and hence the matrix $\m B$ cannot be inverted. Therefore, we propose to redefine the implicit function by adding constraints $\v h(\v\unn)=\v0$ between the model parameters only, \ie, we use the extended implicit function with their Jacobians 
\begin{equation}
 \zvector{\v g(\v y, \v\unn)}{\v h(\vunn)} = \d 0 \eqwith  \m A = \mat{c}{\partial \v g/\partial \v y \\ \m 0\trans} \,, \quad  \m B= \mat{c}{\partial \v g/\partial \v \unn\\ \partial \v h/\partial \v \unn} \,.
\end{equation}
This way we have exactly the same number of constraints as the parameters, and -- except for critical geometric configurations -- the matrix $\m B$ is regular. The following paragraphs list the minimal set of constraints $\v g$, $\v h$ used by individual solvers in our paper.

\subsubsection{Homography estimation.}
The homography matrix has eight degrees of freedom and is defined by the nine parameters $\v\unn$, \ie, the elements of the matrix $\m H \in \mathbb{R}^{3 \times 3}$. In the case of PCs, we used eight constraints $\v g$ from 4 PCs following~\cite{hartley2003multiple} (the DLT algorithm for homography estimation) and one constraint $h(\m H) = ||\vec( \m H)|| - 1 = 0$, which avoids the trivial all-zeros solution. Observe, the this norm constraint influences both, the scale of $\m H$ and the scale of its covariance matrix.

Assuming two ACs, one can select the subset of constraints used for the uncertainty propagation. We defined $\v g$ as four constraints from two point correspondences~\cite{barath2017theory} (Eqn. 1 in \cite{barath2017theory}), and four constraints from one affinity matrix~\cite{barath2017theory} (Eqn. 4 in \cite{barath2017theory}). The constraint $h(\m H) = ||\vec (\m H)|| - 1 = 0$ is the same as for PCs.       

\subsubsection{Fundamental matrix estimation.}
The fundamental matrix has seven degrees of freedom and is defined by the nine parameters of the matrix $\m F \in \mathbb{R}^{3 \times 3}$. In the case of PCs, we used seven constraints $\v g_i = \v y_i^\top \m F \v z_i = 0$ where $i \in \{1,\dots,7\}$ and two constraints $\v h = [h_1, h_2]\trans= \d 0$, where $h_1(\m F) = \det(\m F) - 1 = 0$ and $h_2(\m F) = ||\vec (\m F)|| - 1 = 0$. 

The constraints for 3 ACs are composed of (1) three point constraints following~\cite{hartley2003multiple} of the form
\begin{equation}
    c_{p_i}:=\v z_i\trans \,\m F \,\v y_i=0\,, \quad i=1,2,3\,,
\end{equation}
(2) three pairs of affine constraints~\cite{barath2017minimal}, (Eqn. 8 in \cite{barath2017minimal}) of the form
\begin{equation}
    \underset{2\times 1}{\d c_{a_i}}:= [\m I_2\mid \v 0]\,\m F \,\v y_i+[\m A_i\transi\mid \d 0]\,\m F\trans \,\v z_i=\d 0\,, \quad i=1,2,3\,,
\end{equation}
and (3) the two constraints $\v h = [h_1, h_2]\trans= \d 0$. The propagation function uses the three point constraints, \ie, $c_{p_i}=0$ ($i=1,2,3$), the first two pairs $\d c_{a_i}= \d 0$ ($i=1,2$) of the three pairs of affine constraints, and the two constraints $\v h= \d 0$ on $\m F$.

\subsubsection{Essential matrix estimation.}
We list two ways how to propagate the uncertainty to the parameters of the essential matrix. The essential matrix $\m E \in \mathbb{R}^{3\times3}$ has nine variables and five degrees of freedom. Assuming point correspondences, a straightforward solution is to employ five points constraints  $\v g_i = \v y_i^\top \m E \v z_i = 0$, where $i \in \{1,\dots,5\}$ and four constraints $\v h = [h_1, h_2, h_3, h_4]\trans$ on $\vec (\m E)$. The constraints $h_1(\m E) = \det(\m E) = 0$ and $h_2(\m E) = ||\vec(\m E)||^2 - 2 = 0$ correspond to those for the fundamental matrix. The essential matrix has also nine trace constraints $\m C := 2\m E \m E^\top \m E - \mbox{tr}(\m E \m E^{\top}) \m E = \m 0$. Generally, we can select any two of this nine constraints such that the constraints in $\v h$ are independent, except for certain cases, such as $\m E= [1;0;0]_\times$, where the Jacobian $\partial \vec(\m C)/\partial \vec(\m E)$ has 4 zero rows, not allowing us to use the corresponding constraints, and one is proportional to the Jacobian of the determinant constraint. 

The propagation can also be obtained by using a minimal set of parameters, \eg, a unit translation vector $\v b = [b_1, b_2, b_3]^\top$ and the Euler vector parametrization of the rotation $\m R(\v{r}) \in \mbox {SO(3)}$. For this parametrization, we assume the essential matrix has the form $\m E = [\v b]_x \m R(\v{r})$, where
\begin{equation}
    [\v b]_\times = \mat{ccc}{0 & -b_3 & b_2 \\ b_3 & 0 & -b_1 \\ -b_2 & b_1 & 0} \enspace , \quad \quad \alpha = \sqrt{\v{r}^\top \v{r}} \enspace ,
\end{equation}
\begin{equation}
   \m R(\v r) = \m I_{3} + (1-\cos\alpha) \,[\v r]_\times^2 + \sin\alpha \, [\v r]_\times  \enspace .
\end{equation}
Here, the matrix $\m I_{3} \in \mathbb{R}^{3x3}$ is the identity matrix. This representation leads to five points constraints and only one constraint $h(\v b) = ||\v b|| - 1 = 0$. We used this second representation for our experiments.

The minimal problem using two AC has three constraints for each point, \ie,  $\v g_i = \v y_i^\top \m E \v z_i = 0$, and constraints provided by Eqn. 9 and 10 in~\cite{barath2017minimal}. We used the reduced set of parameters $\v b$, $\v{r}$ with the constraint $h(\v b) = ||\v b|| - 1 = 0$ and five of six constraints on the two ACs, \ie, we suppress one equation of the form of Eq. 10 in~\cite{barath2017minimal} to have the same number of constraints as parameters.

\subsubsection{Essential matrix + focal length estimation.}
\label{sec:propagation-Ef}
The essential matrix with focal length has six degrees of freedom and nine parameters as it can be described as $\m F = \m K^{-\top} \m E \m K^{-1}$, where $\m K = \mbox{Diag}([f,f,1])$ is the calibration matrix with focal length $f$. We assume six point or two affine correspondences as input, which lead to the same constraints as for the essential matrix, \ie, six constraints $g(\v \unn)$. Further, we used the constraints $\v h = [h_1, h_2, h_3]\trans$ where $h_1$, $h_2$ are the same constraints as for the fundamental matrix and $h_3$ corresponds to Eqn. 18 from~\cite{kukelova2017clever}.

\subsection{Preemptive Verification}
\label{sec:preemptive}

\begin{figure}[t!]
  \centering	
	\begin{subfigure}[h]{1.0\columnwidth}
        \includegraphics[width=0.49\textwidth]{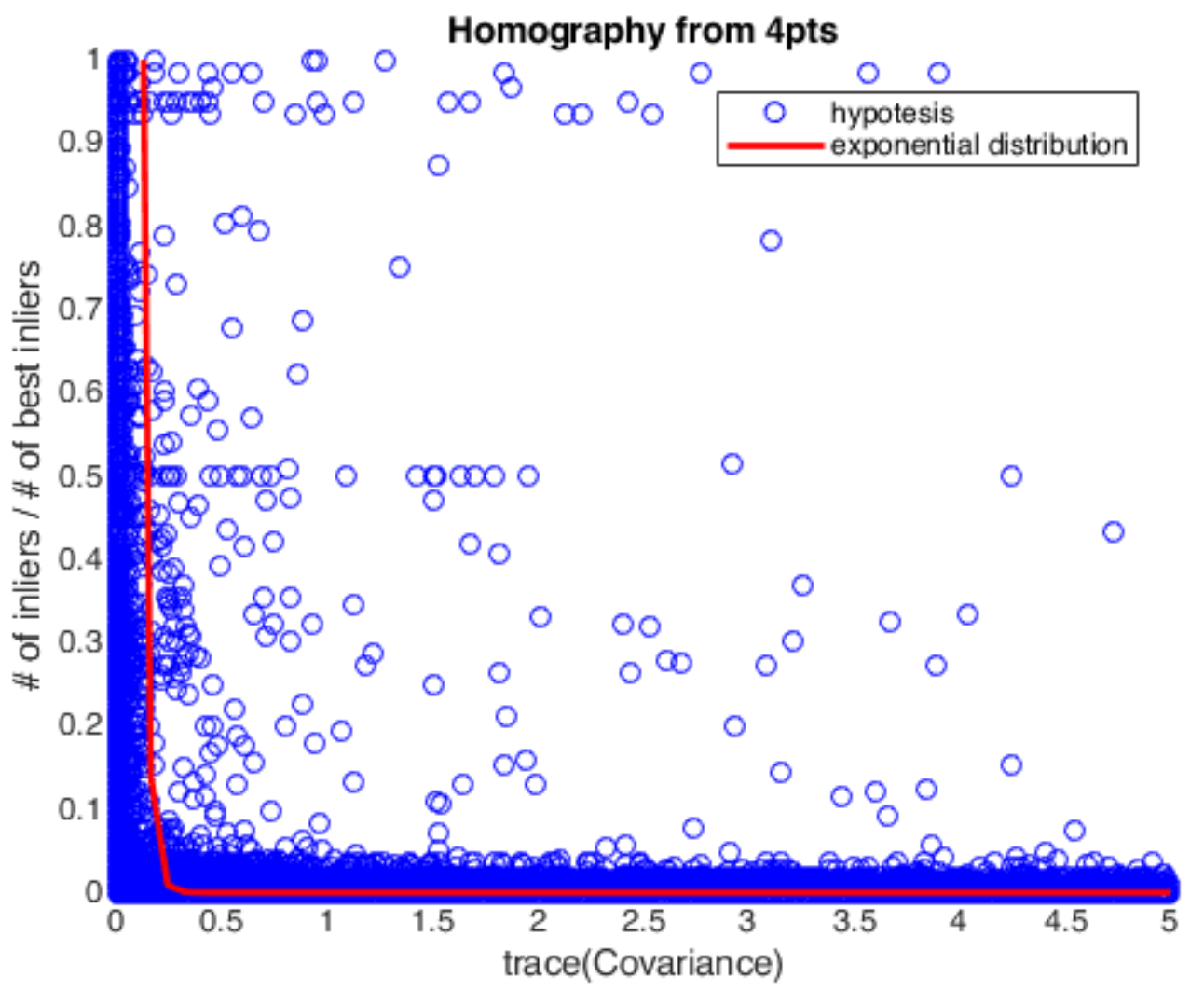}
        \includegraphics[width=0.49\textwidth]{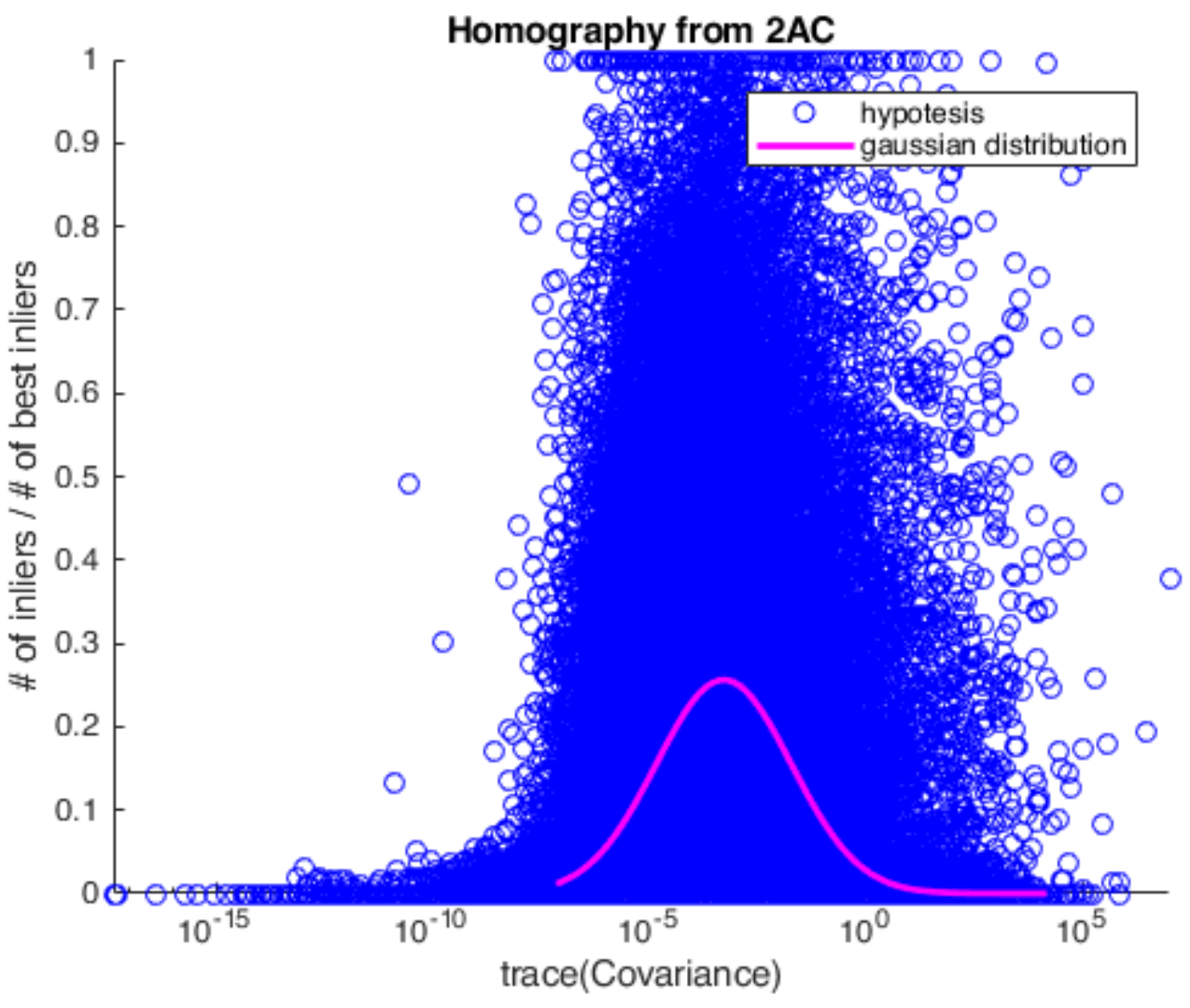}
	    \caption{Homography estimation}
    \end{subfigure}
	\begin{subfigure}[h]{1.0\columnwidth}
        \includegraphics[width=0.49\textwidth]{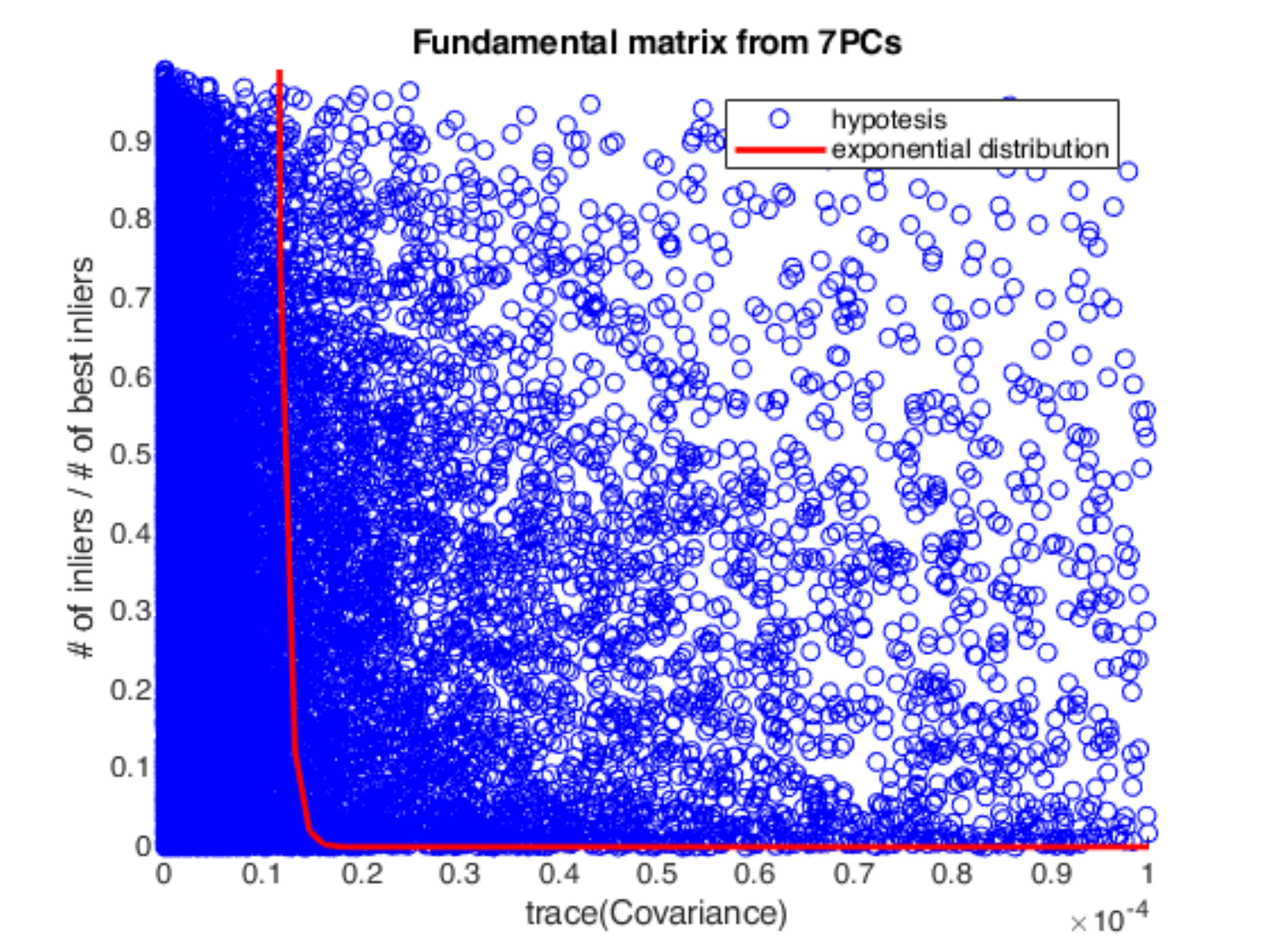}
        \includegraphics[width=0.49\textwidth]{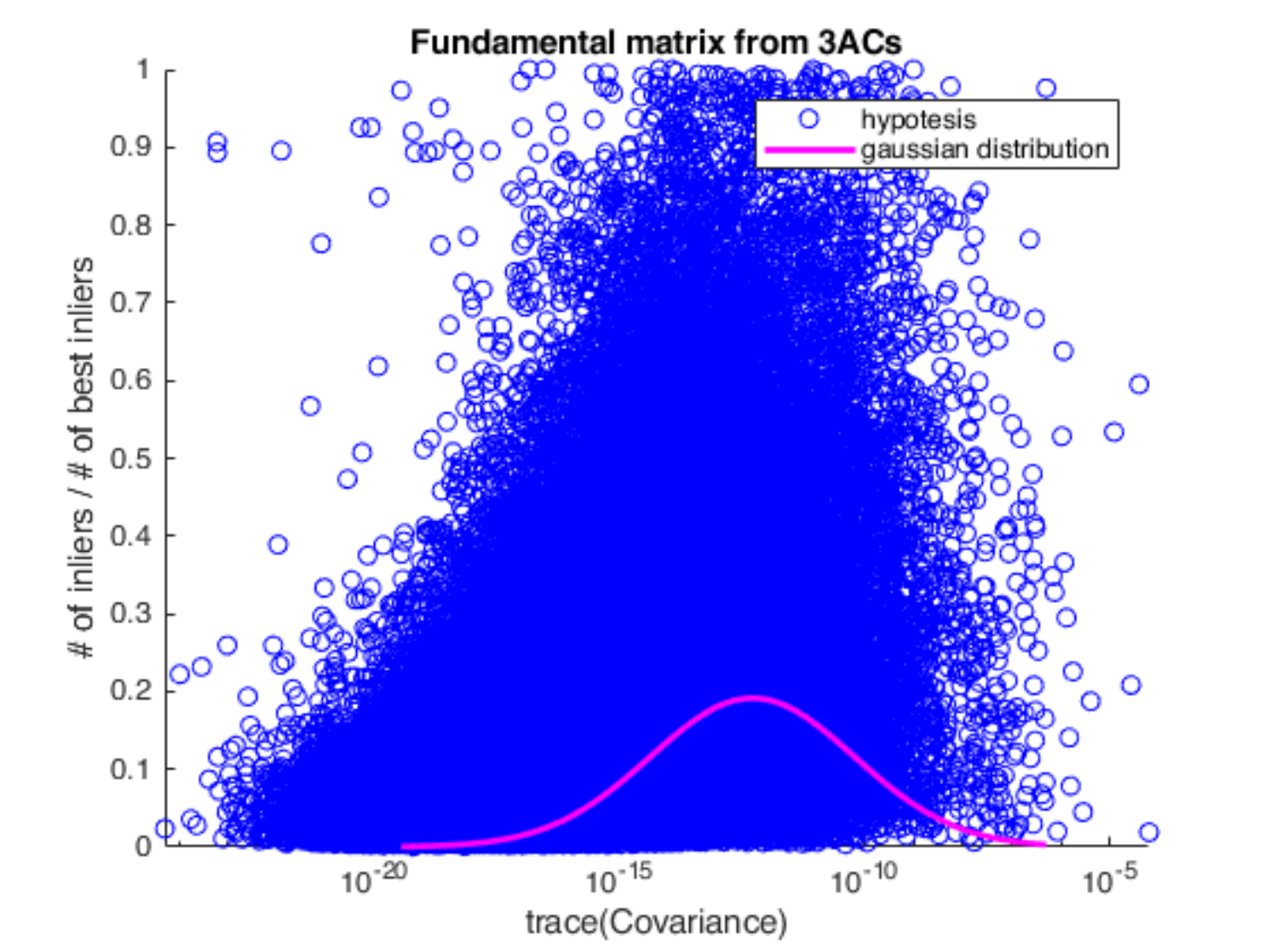}
	    \caption{Fundamental matrix estimation}
    \end{subfigure}
	\caption{  The inlier ratio (vertical) of homographies and fundamental matrices from all tested real scenes as a function of the trace of their covariance matrices (horizontal). This shows that, for points, uncertain models (on right) generate small numbers of inliers. 
	For affine correspondences, the traces have a log-normal distribution. }
	\label{fig:preemption}
\end{figure}

We used the a-priori determined parameters of the inlier ratio distributions (\cf Fig.~\ref{fig:preemption}), to measure the probability of having a {\em good model } based on its uncertainty. 
For determining the parameters of the distributions, we saved the uncertainty (measured by the trace of the covariance matrix) of all models generated in the RANSAC loop, on all scenes from all datasets, and, also, their number of inliers divided by the number of inliers of the best model found on that particular image pair, \ie, we saved their inlier ratios.
We found that calculating the parameters of the distributions using only those models which lead to an inlier ratio of, at least, $0.95$ works well for for all problems considered in the paper. 
%

In our experiments, we fit the exponential distribution parameter $\hat{\lambda} = (n-2) / \sum_{i = 1}^n (t_i)$, where $n$ is the number of good models $\unn_i$ and $t_i$ is the trace of covariance matrices, \ie, $t_i = \mbox{tr}(\Sigma_{\unn_i \unn_i})$. 
For solvers using affine correspondences, the distributions are described by a log-normal distribution, \ie, by mean $\mu = 1/n \sum_{i = 1}^n \log_{10}(t_i)$ and variance $\sigma^2 = 1/(n-1) \sum_{i = 1}^n (\log_{10}(t_i) - \mu)^2)$.  


\subsection{Local Optimization}

\begin{figure}[t]
  \centering	
	\begin{subfigure}[h]{1.0\columnwidth}
        \includegraphics[width=0.49\textwidth]{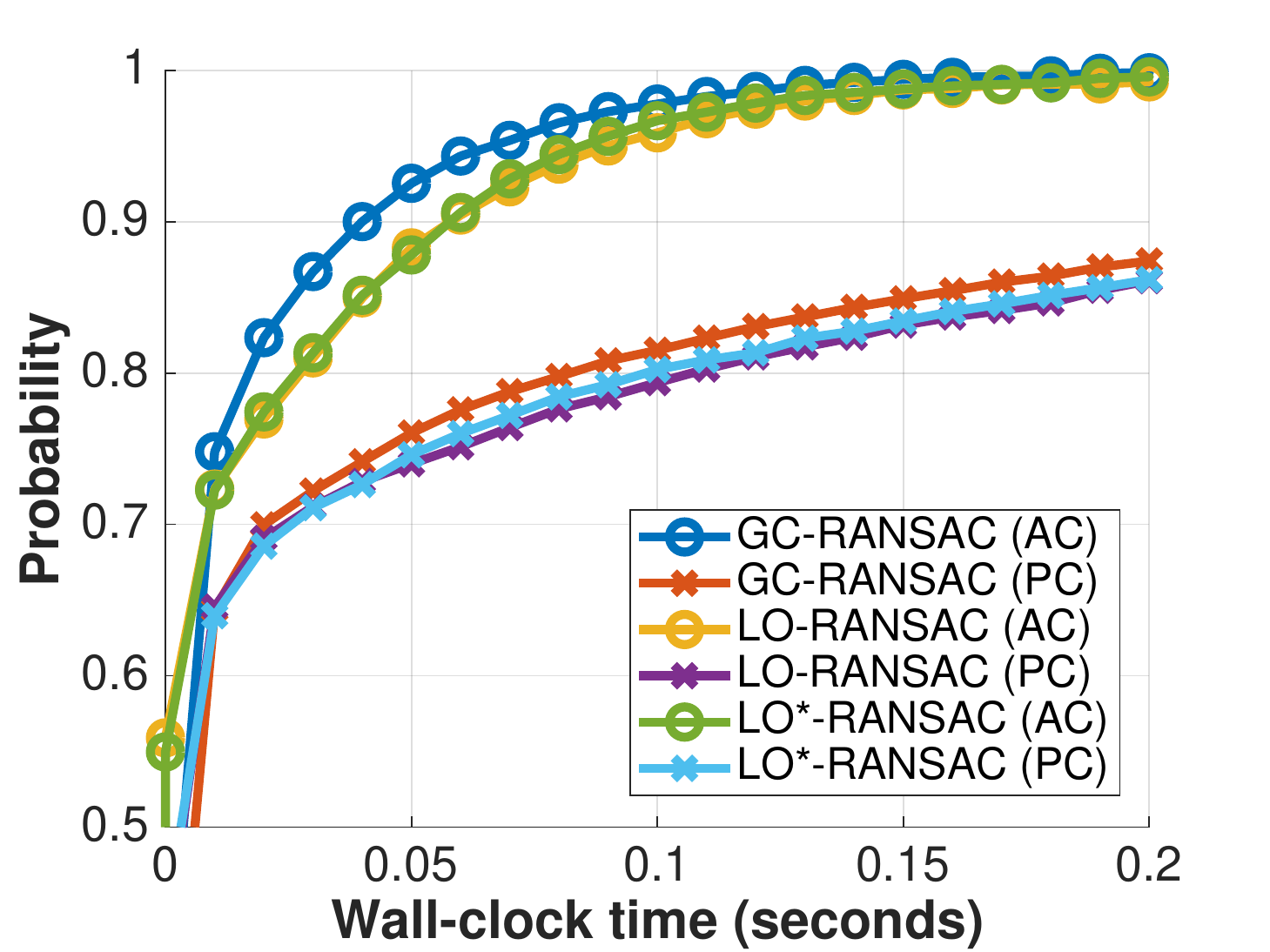}
        \includegraphics[width=0.49\textwidth]{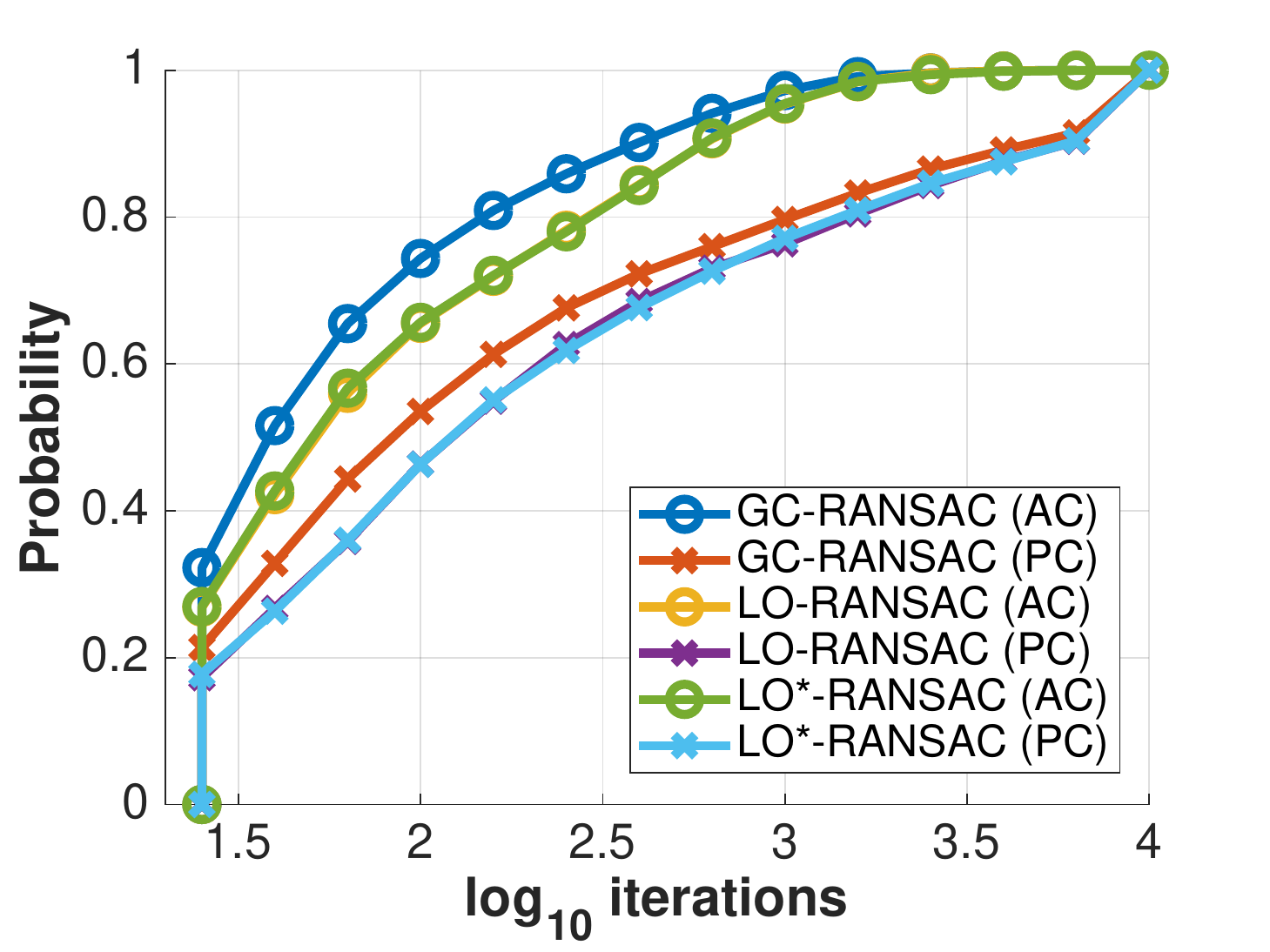}
	    \caption{Homography estimation}
    \end{subfigure}
    \begin{subfigure}[h]{1.0\columnwidth}
        \includegraphics[width=0.49\textwidth]{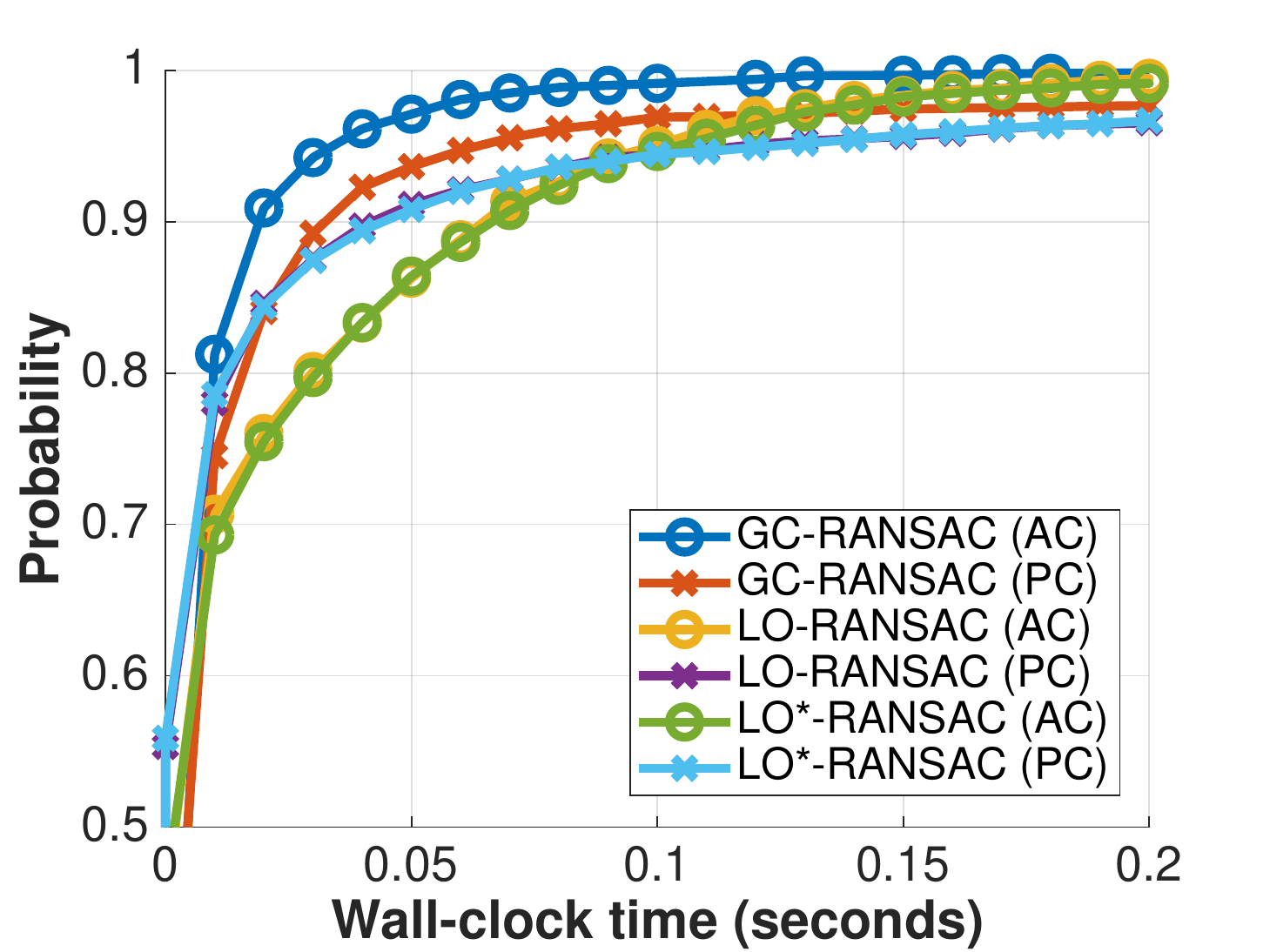}
        \includegraphics[width=0.49\textwidth]{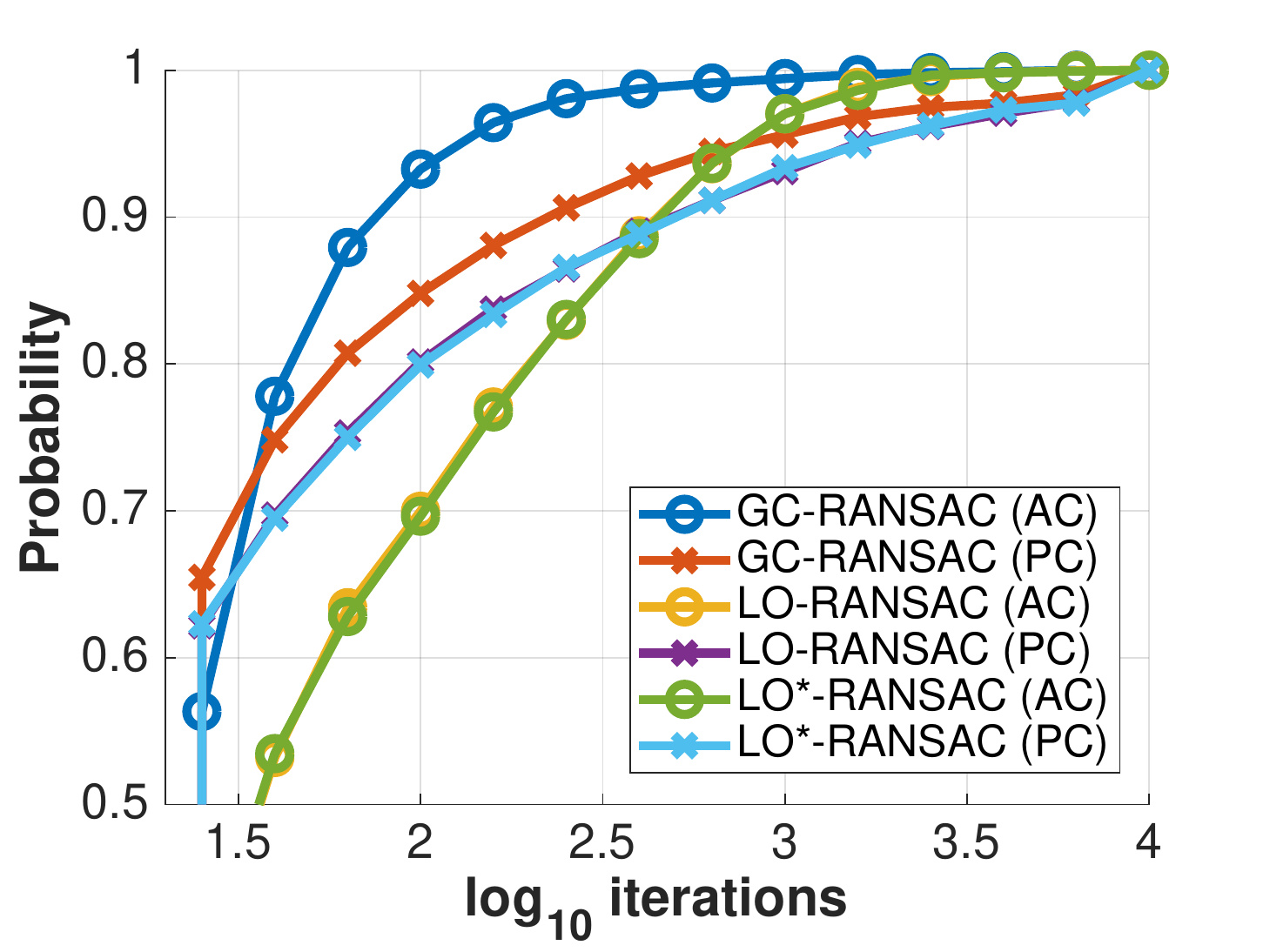}
	    \caption{Fundamental matrix estimation}
    \end{subfigure}
    \begin{subfigure}[h]{1.0\columnwidth}
        \includegraphics[width=0.49\textwidth]{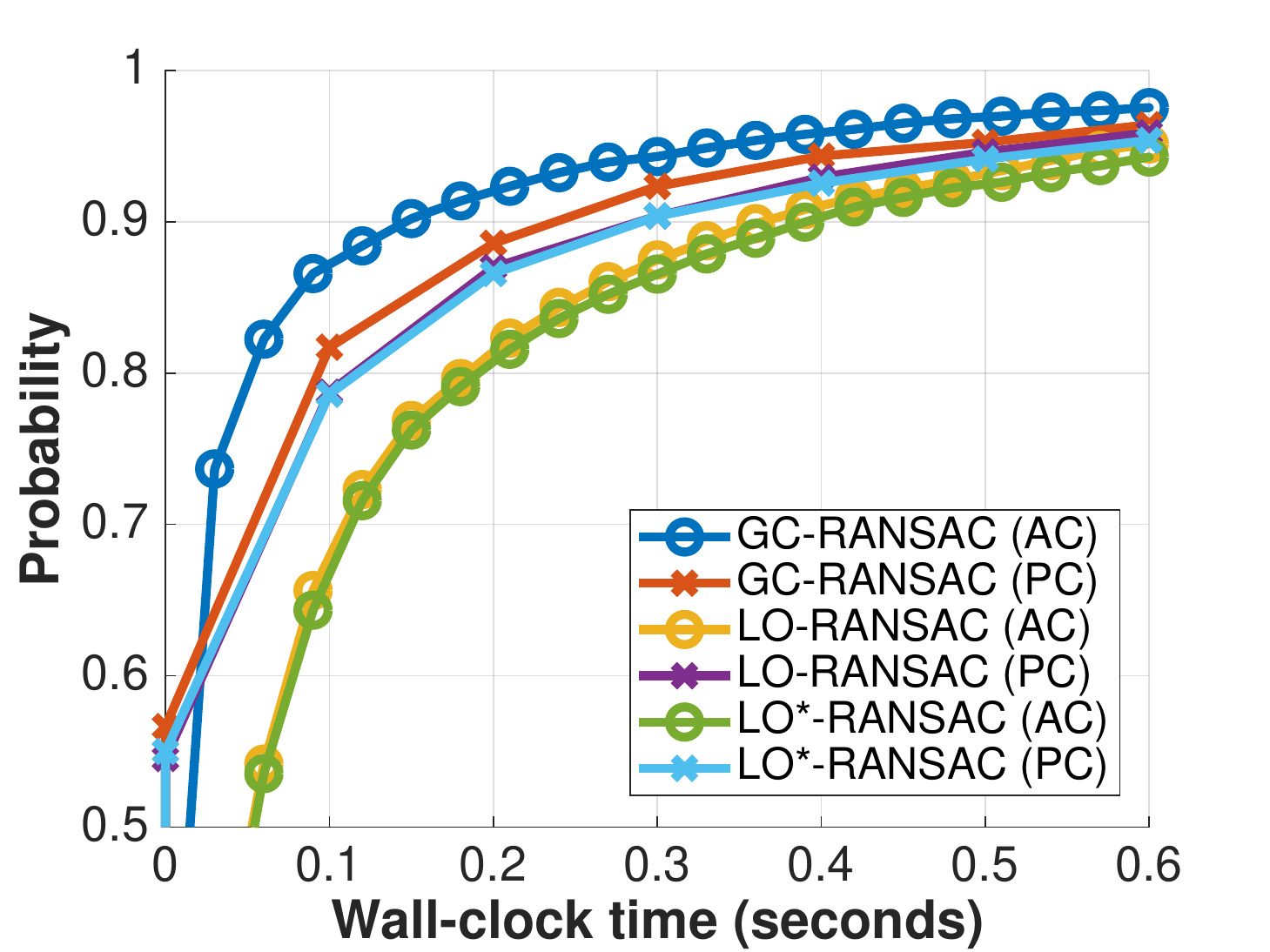}
        \includegraphics[width=0.49\textwidth]{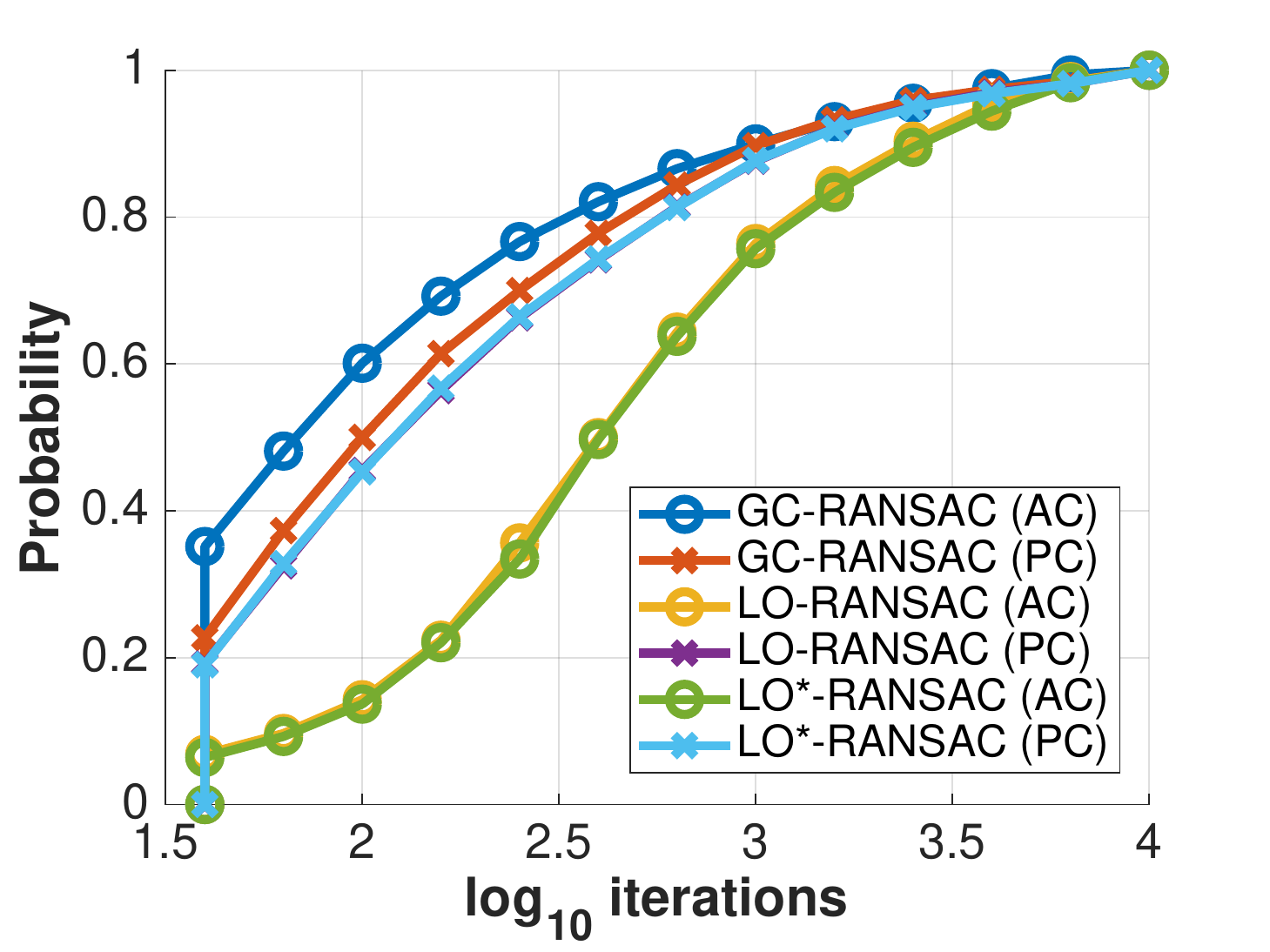}
	    \caption{Essential matrix estimation}
    \end{subfigure}
	\caption{The cumulative distribution functions of the wall-clock times and $\log_{10}$ iteration numbers using different local optimizations for affine correspondence (AC) and point-based (PC) solvers.}
	\label{fig:local-optimization-time}
\end{figure}

The cumulative distribution functions of the wall-clock times and $\log_{10}$ iteration numbers using different local optimizations for affine correspondence (AC) and point-based (PC) solvers are shown in Fig.~\ref{fig:local-optimization-time}. 
For homography and fundamental matrix estimation, the proposed combined preemptive model verification strategy was used. For essential matrices, SPRT~\cite{chum2008optimal} was applied.
It can be seen that Graph-Cut RANSAC~\cite{barath2018graph} (GC-RANSAC) is the fastest and leads to the fewest RANSAC iterations on all investigated problems.
Also, methods using ACs are significantly faster than point-based ones.

\clearpage
\bibliographystyle{splncs04}
\bibliography{egbib,torsten}
\end{document}